\documentclass{article}
\usepackage[T1]{fontenc}
\usepackage{iclr2027_conference,times}

\usepackage{graphicx}
\usepackage{xcolor}
\usepackage{colortbl}
\usepackage{wrapfig}
\usepackage{needspace}

\usepackage{amsmath,amsfonts,bm}

\def\eqref#1{equation~\ref{#1}}

\def\1{\bm{1}}

\DeclareMathAlphabet{\mathsfit}{\encodingdefault}{\sfdefault}{m}{sl}
\SetMathAlphabet{\mathsfit}{bold}{\encodingdefault}{\sfdefault}{bx}{n}

\usepackage{hyperref}
\usepackage{cleveref}

\usepackage{pifont}
\usepackage{booktabs}
\usepackage{array}
\usepackage{tabularx}
\usepackage{multirow}
\usepackage{capt-of}

\usepackage{algorithm}
\usepackage{algpseudocode}
\usepackage{enumitem}

\definecolor{rowblue}{RGB}{220,240,248}
\newcolumntype{Y}{>{\centering\arraybackslash}X}

\definecolor{BDPink}{HTML}{D41472}
\definecolor{BDBlue}{HTML}{008AC4}
\definecolor{BDRow}{HTML}{E5F3FA}
\definecolor{BDVenue}{HTML}{555555}

\newcommand{\bddatasetfinal}[1]{{\footnotesize\bfseries #1}}
\newcommand{\bdfirstfinal}[2]{\textcolor{BDPink}{$\mathbf{#1}^{#2}$}}
\newcommand{\bdsecondfinal}[2]{\textcolor{BDBlue}{$\mathbf{#1}^{#2}$}}

\newcommand{\cmark}{\ding{51}}
\newcommand{\xmark}{\ding{55}}

\title{ResARC: Residual-Aware AutoRegressive Coding for Ultra-Low Bitrate Image Compression}

\author{\textbf{Qin Yan$^{1}$\thanks{Equal contribution.}\quad Ruixiao Dong$^{1}$\footnotemark[1]\quad Yutao Xie$^{1}$\quad Li Li$^{1}$\thanks{Corresponding author.}\quad Ying Chen$^{2}$\quad Kai Li$^{2}$}\\
\textbf{Daowen Li$^{2}$\quad Houqiang Li$^{1}$}\\
{\normalfont\small
$^{1}$University of Science and Technology of China}
\\
{\normalfont\small
$^{2}$Alibaba Group}
\\
{\normalfont\footnotesize\texttt{
\{yanqin1,dongruixiaoyx,yutaoxie\}@mail.ustc.edu.cn}}
\\
{\normalfont\footnotesize\texttt{
\{lilimao,lihq\}@ustc.edu.cn}}
\\
{\normalfont\footnotesize\texttt{
\{chenying.ailab,kaishi.lk,lidaowen.ldw\}@alibaba-inc.com}}
}

\definecolor{resarcbestcell}{RGB}{255,220,230}
\definecolor{resarcsecondcell}{RGB}{221,235,247}

\definecolor{resarcquestionbg}{RGB}{238,245,252}
\definecolor{resarcquestionborder}{RGB}{83,124,165}
\definecolor{resarcquestiontext}{RGB}{28,65,103}

\renewcommand{\topfraction}{0.90}
\renewcommand{\dbltopfraction}{0.90}
\renewcommand{\textfraction}{0.10}
\iclrfinalcopy

\begin{document}
\raggedbottom

\maketitle
\lhead{Preprint}

\begin{figure*}[h]
  \centering
  \vspace{-1.0cm}
  \includegraphics[width=1.0\textwidth]{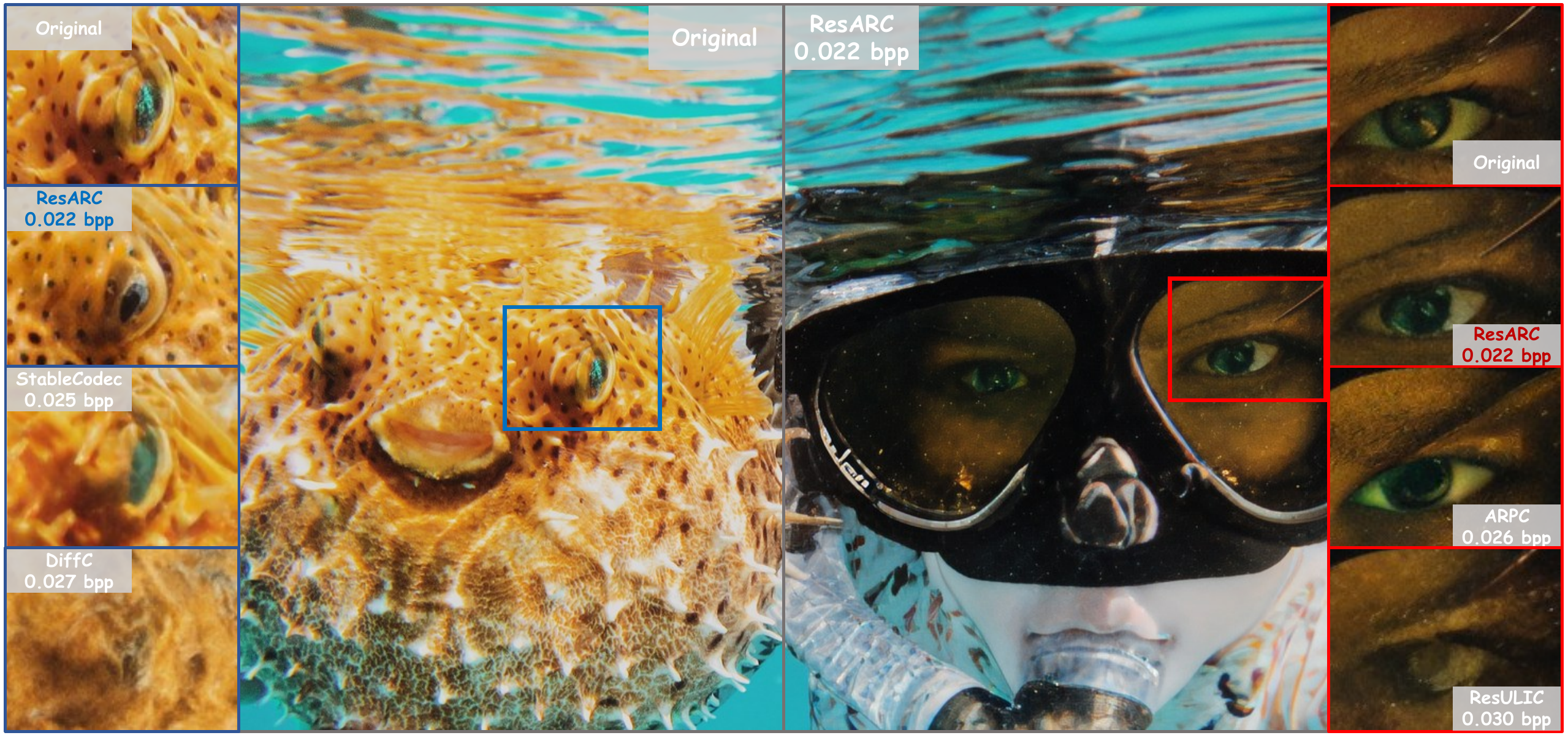}
  \vspace{-0.5cm}
  \caption{\textbf{Qualitative comparison with baselines at comparable ultra-low bitrates.} 
  ResARC better preserves fine-grained textures and image structures than competing generative codecs.}
  \vspace{-0.1cm}
  \label{fig:teaser}
\end{figure*}

\begin{abstract}

Progressive autoregressive image codecs provide an appealing paradigm for generative compression by quantizing continuous latents into discrete tokens, transmitting coarse-to-fine prefix tokens and generating the remaining suffix tokens at the decoder.
However, their reconstruction quality is fundamentally limited by two residuals introduced along this pipeline: the \textit{quantization residual}, arising from information loss during discrete tokenization, and the \textit{generation residual}, resulting from imperfect autoregressive generation of the suffix tokens.
To address these limitations, we introduce \textbf{ResARC}, a residual-aware autoregressive codec that explicitly compensates for both residuals at the decoder.
Specifically, we generate the quantization residual with a diffusion transformer conditioned on the autoregressive decoding context, while requiring no additional side information.
In parallel, we compute the generation residual at the encoder and employ a learned Generation Residual Codec to efficiently compress and transmit it for decoder-side correction.
The recovered residuals are then integrated with the reconstructed latent representation and decoded through an adapted VAE decoder.
Extensive experiments demonstrate that ResARC achieves competitive perceptual similarity while substantially improving distributional fidelity over leading generative codecs across the ultra-low bitrate regime. Code and models will be released soon.

\end{abstract}

\section{Introduction}

\begin{wrapfigure}{R}{0.50\textwidth}
  \centering
  \vspace{-0.3cm}
  \includegraphics[width=1.0\linewidth]{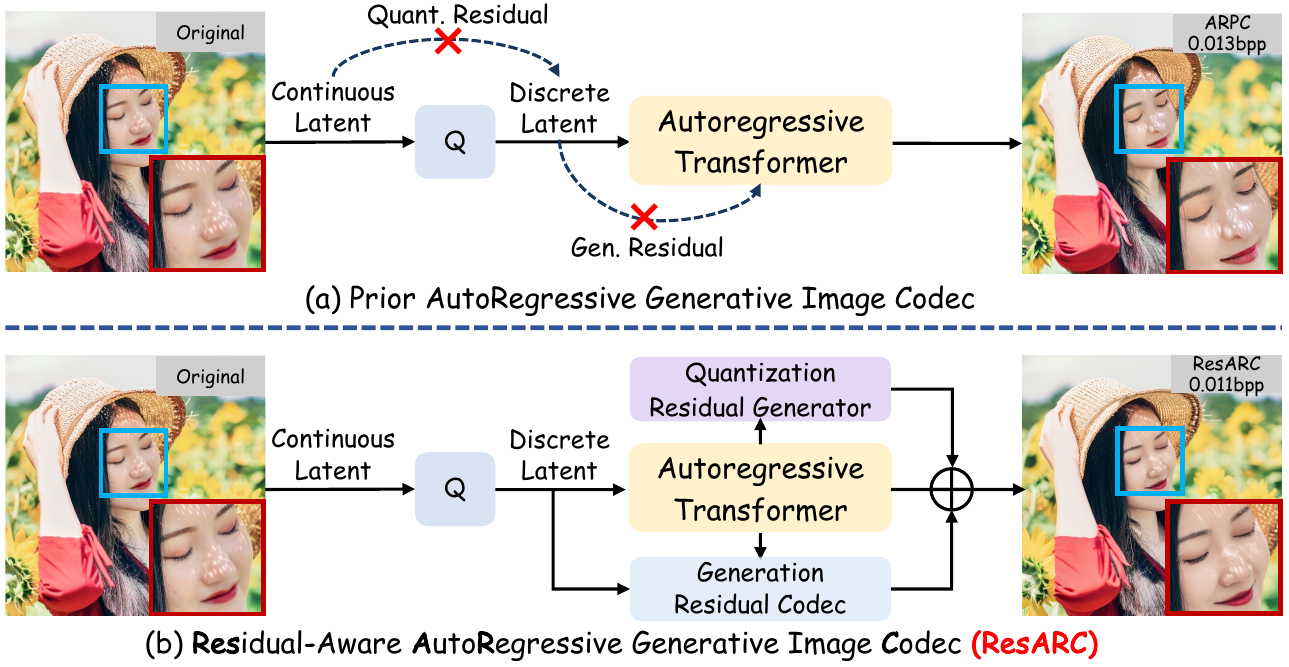}
  
  \caption{\textbf{Residual-aware design of ResARC.}
(a) Prior autoregressive codecs introduce \emph{quantization and generation residuals}.
(b) ResARC generates the former and transmits the latter.}
  
  \label{fig:motivation}
\end{wrapfigure}
The rapid growth of visual data has made efficient image compression increasingly important, particularly under limited bandwidth and storage.
At ultra-low bitrates, the limited transmitted information makes it difficult to preserve fine textures and image structures, often leading distortion-oriented codecs~\citep{balle2017endtoend,balle2018hyperprior} to produce overly smooth reconstructions.
Generative codecs~\citep{agustsson2019gancompression,mentzer2020hific,careil2024perco} address this by leveraging generative priors to synthesize plausible details from compressed representations, substantially improving perceptual quality at low bitrates.
Early generative codecs mainly rely on GAN priors~\citep{agustsson2019gancompression,mentzer2020hific}, and more recent approaches adopt diffusion models for high-quality generative reconstruction~\citep{theis2022diffc}.
However, these approaches generally lack native progressive bitrate control and tight integration between entropy modeling and the generative prior.

Autoregressive generative codecs provide an alternative by exploiting the shared autoregressive prior for both entropy modeling and token generation~\citep{mao2024extreme,xue2024unifying}.
More recently, Visual Auto-Regressive (VAR) models~\citep{tian2024var,han2025infinity} generate visual tokens in a coarse-to-fine manner across multiple scales, providing a natural foundation for progressive generative compression.
Building on this paradigm, ARPC~\citep{zhang2026arpc} quantizes continuous latents into discrete tokens, transmits a coarse-to-fine prefix of tokens, and autoregressively generates the remaining suffix at the decoder.
By varying the transmitted prefix depth, a single model can operate at multiple bitrates.

Despite their scalability and flexibility, we observe that the reconstruction fidelity of autoregressive generative codecs remains limited by two residuals arising along the pipeline.
As illustrated in~\Cref{fig:motivation}, discrete tokenization converts the continuous latent into transmissible tokens, but inevitably discards information, including fine-grained details important for reconstruction.
We define the resulting difference between the original continuous latent and its quantized reconstruction as the \emph{quantization residual}.
A second discrepancy arises when the suffix tokens omitted from transmission are replaced by autoregressively synthesized predictions, which may deviate from their ground-truth counterparts.
We define this discrepancy as the \emph{generation residual}, which further degrades reconstruction fidelity.
These observations raise a central question:
\textcolor{blue}{
\textit{\textbf{How should quantization and generation residuals be explicitly compensated in autoregressive compression?}}}

To answer this question, we introduce \textbf{ResARC}, the first residual-aware autoregressive image codec that explicitly compensates for both residuals according to their distinct characteristics.
Specifically, the \emph{quantization residual} is predominantly associated with fine-grained details discarded during discrete tokenization, while the reconstructed tokens and autoregressive decoding context available at the decoder provide strong cues for recovering this missing information.
We therefore employ a \textbf{Quantization Residual Generator}, which uses a conditional diffusion transformer to generate the quantization residual from the decoding context without transmitting additional side information.
For the \emph{generation residual}, the discrepancy between the ground-truth and generated suffix may affect both structural content and fine details.
Since this residual represents the specific correction required to align the generated suffix with its ground-truth counterpart, we compute it at the encoder and employ a \textbf{Generation Residual Codec} to efficiently compress it into a bitstream transmitted for decoder-side correction.
Finally, the recovered quantization and generation residuals are fused with the reconstructed latent and decoded via an adapted VAE decoder, improving reconstruction fidelity while preserving progressive bitrate control.

Quantitative evaluations on DIV2K~\citep{agustsson2017div2k} and CLIC2020~\citep{toderici2020clic} demonstrate that ResARC achieves strong perceptual similarity and substantially improved distributional fidelity compared with leading autoregressive and diffusion-based generative codecs in the ultra-low bitrate regime.
Qualitative comparisons further show that ResARC better preserves fine-grained textures and image structures at comparable or lower bitrates.
Moreover, ablation studies validate the benefits of both residual branches and the effectiveness of their respective compensation strategies.

Our principal contributions are summarized as follows:
\begin{itemize}[leftmargin=10pt, itemsep=2pt, topsep=0pt, parsep=0pt]

\item We identify two distinct residuals in autoregressive generative codecs: the \emph{quantization residual} introduced by discrete tokenization, and the \emph{generation residual} arising from imperfect autoregressive suffix generation.
\item We propose \textbf{ResARC}, the first residual-aware autoregressive codec that generates the quantization residual without additional transmitted bits and compresses the generation residual into a compact bitstream transmitted for decoder-side correction.
\item Extensive experiments on DIV2K and CLIC2020 demonstrate improved perceptual similarity and distributional fidelity in the ultra-low bitrate regime, with ablations further validating the complementary contributions of the two residual branches.
\end{itemize}

\begingroup
\setlength{\parskip}{3pt}
\setlength{\abovedisplayskip}{5pt plus 1pt minus 1pt}
\setlength{\belowdisplayskip}{5pt plus 1pt minus 1pt}
\setlength{\abovedisplayshortskip}{0pt plus 1pt}
\setlength{\belowdisplayshortskip}{3pt plus 1pt minus 1pt}

\section{Related Work}

\label{sec:related_work}
\paragraph{Generative image compression.} 
Image compression has traditionally relied on hand-crafted transforms and coding tools~\citep{BPG}, whereas modern neural codecs~\citep{balle2017endtoend} jointly learn representations, entropy models, and reconstruction in an end-to-end manner.
While these methods primarily optimize the conventional rate-distortion trade-off, producing perceptually realistic reconstructions remains challenging, particularly in the ultra-low bitrate regime.
Recent advances in visual generative modeling~\citep{rombach2022high} have therefore motivated generative codecs that leverage learned generative priors to improve perceptual reconstruction quality.
GAN-based codecs established this paradigm for low bitrate perceptual compression~\citep{agustsson2019gancompression,mentzer2020hific}, with later approaches further improving statistical fidelity~\citep{muckley2023msillm}.
More recently, diffusion-based codecs have conditioned generation on compressed representations~\citep{yang2023cdc,hoogeboom2023hfd}, semantic or visual cues~\citep{careil2024perco,korber2024percosd}, and semantic-space residuals~\citep{ke2025resulic}.
Subsequent works have improved efficiency and flexibility through shortened or one-step sampling~\citep{relic2024foundation,zhang2025stablecodec,xue2025onedc,shi2026ditic}, multi-rate coding~\citep{guo2025oscar}, lightweight architectures~\citep{zhang2026aeic,jia2026codlite}, compression-oriented pretraining~\citep{jia2026cod}, video-diffusion decoding~\citep{chen2026nefic}, and content-adaptive coding~\citep{sheng2026cadc}.
Beyond diffusion-based approaches, GLC~\citep{jia2024glc} and DLF~\citep{xue2025dlf} explore generative latent compression, while RDVQ~\citep{jiang2026rdvq} improves rate-distortion optimization for vector-quantized representations.
Despite these advances, existing generative codecs generally lack native progressive bitrate control and tight integration between entropy modeling and the generative prior.

\paragraph{Autoregressive generative codecs.} 
Autoregressive modeling over discrete visual tokens~\citep{oord2017vqvae,esser2021vqgan,lee2022rqtransformer,yu2024magvit2,zhao2025bsq} provides a natural way to address the above limitations by sharing a generative prior for both entropy modeling and sequential generation, and the autoregressive prior further enables progressive coding.
Early autoregressive codecs~\citep{mao2024extreme,xue2024unifying} explored this direction with conventional next-token prediction.
More recently, Visual Auto-Regressive (VAR) models~\citep{tian2024var,han2025infinity} generate token maps in a coarse-to-fine hierarchy, providing a natural basis for progressive generative compression, while HART~\citep{tang2025hart} further enhances this paradigm with continuous residual diffusion to improve image synthesis.
Building on this coarse-to-fine autoregressive generation, \cite{phung2025exploring} first explored its application to image compression through preliminary experiments.
ARPC~\citep{zhang2026arpc} further develops this paradigm by transmitting coarse-to-fine prefix tokens and autoregressively generating the remaining suffix tokens, with the transmitted prefix depth controlling the bitrate within a single model.
ProGVC~\citep{li2026progvc} extends progressive autoregressive coding to generative video compression.
Despite these advances, existing autoregressive codecs remain limited by two distinct residuals: discrete tokenization discards information from continuous latents, yielding a \emph{quantization residual}, while imperfect autoregressive suffix generation introduces a \emph{generation residual}. ResARC explicitly models and compensates for both residuals to improve reconstruction fidelity.

\section{Background: Multi-Scale Autoregressive Image Compression}

\label{sec:preliminaries}
We first review the progressive autoregressive codec ARPC~\citep{zhang2026arpc} based on Infinity~\citep{han2025infinity}, which underlies our method.
Given an input image $x\in\mathbb{R}^{H\times W\times 3}$, a VAE encoder first extracts a continuous latent representation $h=\mathcal{E}(x)$. 
A multi-scale residual quantizer~\citep{zhao2025bsq} $\mathcal{Q}$ then converts $h$ into a sequence of discrete token maps $\mathcal{T}=(T_1,\ldots,T_K)=\mathcal{Q}(h)$, where successive scales progressively refine the latent representation from coarse structures to fine details. 
The complete source token sequence is aggregated through multi-scale summation to obtain the quantized latent $h_q=\mathcal{S}(\mathcal{T})=\sum_{i=1}^{K}U_i(T_i)$.
Here, $U_i$ maps the scale-$i$ tokens to latent features at the common spatial resolution of $h$.
For a given prefix depth $k$, the codec transmits the first $k$ token scales $T_{\leq k}$ and leaves the remaining suffix tokens $T_{> k}$ to be synthesized at the decoder.
The autoregressive model $\psi$ is utilized for both entropy coding and suffix generation: its predicted probabilities are used to entropy-code the transmitted prefix and to autoregressively sample the omitted suffix scales.

Specifically, given the text condition $\ell$, the autoregressive model estimates the prefix token bitrate in bits per pixel (bpp) as

\begin{equation}
R_{\mathrm{prefix}}^{(k)}=-\frac{1}{HW}\sum_{i=1}^{k}\log_2 p_\psi(T_i\mid T_{<i},\ell).
\end{equation}
After entropy decoding the prefix scales from the bitstream, the same autoregressive transformer predicts the conditional distribution for each remaining suffix scale and samples the corresponding tokens as

\begin{equation}
    \hat{T}_i \sim p_{\psi}\left( T_i \mid T_{\leq k}, \hat{T}_{k+1:i-1}, \ell \right), \qquad i=k+1,\ldots,K.
\end{equation}
The transmitted prefix and generated suffix are then combined to form the complete multi-scale token sequence $\hat{\mathcal{T}}^{(k)}=(T_1,\ldots, T_k,\hat{T}_{k+1},\ldots,\hat{T}_K)$, which is aggregated and sent to the decoder for reconstruction:

\begin{equation}
\hat h_q^{(k)}=\mathcal S\left(\hat{\mathcal T}^{(k)}\right),
\qquad
\hat x^{(k)}=\mathcal D\left(\hat h_q^{(k)}\right).
\end{equation}
By varying the prefix depth $k$, the codec changes the number of transmitted token scales and therefore supports progressive bitrate control within a single model.

This progressive coding paradigm reconstructs images from quantized token representations and autoregressively generated suffixes, naturally introducing two distinct residuals that motivate the residual-aware design of ResARC.

\section{ResARC: Residual-aware Autoregressive Codec}

\label{sec:method}

\begin{figure*}[!t]
  \centering
  
  \includegraphics[width=1.0\linewidth]{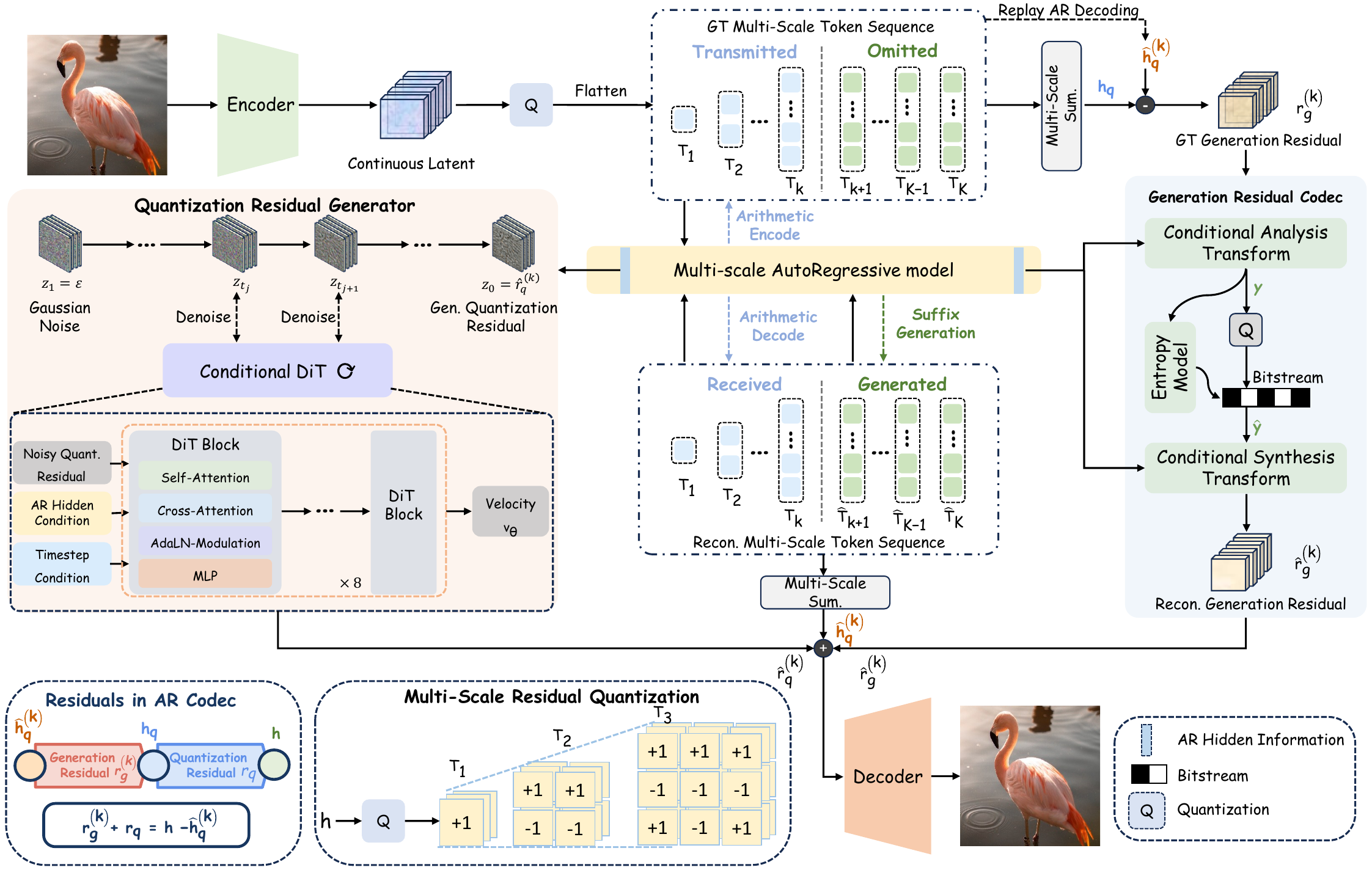}
  
  \caption{\textbf{Overview of the ResARC architecture.} 
  ResARC decomposes the discrepancy in the autoregressive codec into a quantization residual and a generation residual.
  The quantization residual is synthesized at the decoder without additional bits, while the generation residual is compressed and transmitted for decoder-side correction.
}

  \label{fig:resarc_overview}
\end{figure*}
\subsection{Residual-Aware Formulation}

ResARC builds on the progressive autoregressive coding framework introduced above.
As discussed, this coding pipeline introduces two distinct discrepancies.
The first arises from discrete tokenization.
To obtain transmissible discrete symbols, the continuous latent $h$ extracted by the VAE encoder is quantized into multi-scale token maps, which are aggregated to form the quantized latent $h_q$.
Since discrete tokenization cannot preserve all information in $h$, the quantized latent $h_q$ generally differs from the original continuous latent.
We define the resulting discrepancy as the \emph{quantization residual}:
\begin{equation}
r_q=h-h_q=h-\mathcal{S}\left(\mathcal{Q}(h)\right).
\end{equation}
This residual represents information lost during tokenization, including fine-grained details that are vital for reconstruction.

The second discrepancy arises from autoregressive suffix generation.
For a prefix depth $k$, only the prefix scales $T_{\leq k}$ are transmitted, while the omitted suffix scales are autoregressively synthesized as $\hat T_{>k}$ at the decoder.
Aggregating the decoded prefix and generated suffix yields the reconstructed latent
\begin{equation}
\hat h_q^{(k)}=\mathcal S\left(\hat{\mathcal T}^{(k)}\right)=\mathcal S\left(T_{\leq k},\hat T_{>k}\right).
\end{equation}
Since the generated suffix may deviate from its ground-truth counterpart, $\hat h_q^{(k)}$ can differ from the quantized latent $h_q$ reconstructed from the complete ground-truth token sequence.
We define this discrepancy as the \emph{generation residual}:
\begin{equation}
\label{eq:def_gen_res}
r_g^{(k)}=h_q-\hat h_q^{(k)}=\mathcal S(\mathcal T)-\mathcal S\left(\hat{\mathcal T}^{(k)}\right).
\end{equation}
This generation residual captures deviations introduced by autoregressive suffix generation and may affect both local details and image structures.
Overall, the discrepancy between the continuous latent $h$ and the autoregressively reconstructed latent $\hat h_q^{(k)}$ decomposes into the two residuals:
\begin{equation}
\label{eq:residual_decomposition}
h-\hat h_q^{(k)}
=
(h-h_q)+(h_q-\hat h_q^{(k)})
=
r_q+r_g^{(k)}.
\end{equation}
This decomposition motivates ResARC to compensate for the two residuals with dedicated strategies: generating the quantization residual from decoder-side context and explicitly transmitting a compact correction for the generation residual.

\subsection{Quantization Residual Generation}
\label{sec:quant_residual_generation}

The \textit{quantization residual} $r_q$ represents information lost when the continuous latent is quantized into discrete multi-scale token maps.
The reconstructed tokens and autoregressive decoding features available at the decoder provide rich cues for estimating this residual.
We therefore generate the residual at the decoder without transmitting additional bits.
To this end, we introduce the \textbf{Quantization Residual Generator}, implemented as a conditional Diffusion Transformer (DiT)~\citep{peebles2023dit}.
As illustrated in~\Cref{fig:resarc_overview}, the DiT conditioned on hidden features produced during autoregressive decoding iteratively generates the quantization residual from Gaussian noise.
The autoregressive features are injected through cross-attention, while the flow matching timestep modulates the DiT blocks via Adaptive LayerNorm (AdaLN).

To learn this conditional generation process, we train the conditional DiT using flow matching~\citep{lipman2023flowmatching,liu2023flow}.
Let $z_0=r_q$ denote the target residual and $z_1=\epsilon$ denote Gaussian noise, where $\epsilon\sim\mathcal N(0,I)$.
For $t\sim\mathcal U(0,1)$, we define the linear interpolation $z_t=(1-t)z_0+t z_1$ whose target velocity is $v=z_1-z_0$; the training objective is
\begin{equation}
\mathcal L_{\mathrm{FM}}
=
\mathbb E\!\left[
\left\|v_\theta(z_t,t;c_{\mathrm{quant}})-v\right\|_2^2
\right].
\label{eq:quant_flow_matching}
\end{equation}
Here, $c_{\mathrm{quant}}$ denotes the autoregressive decoding features obtained at prefix depth $k$.
At inference, starting from Gaussian noise at $t=1$, we integrate the learned velocity field toward $t=0$ to obtain the predicted quantization residual $\hat r_q^{(k)}=z_0$.

Since the generator relies only on autoregressive features already available in the decoder, quantization residual generation does not require additional transmitted payload.

\subsection{Generation Residual Coding}

\label{sec:generation_residual_coding}

The \emph{generation residual} captures the mismatch between the generated suffix and the ground-truth token sequence.
The decoder can reproduce the generated suffix, but the corresponding ground-truth suffix tokens are only available at the encoder.
We therefore compute the generation residual at the encoder and transmit a compact correction for decoder-side recovery.

Specifically, as illustrated in~\Cref{fig:resarc_overview},
the encoder replays the same autoregressive suffix-generation process as the decoder to reproduce $\hat h_q^{(k)}$, and computes the generation residual $r_g^{(k)}$ according to~\Cref{eq:def_gen_res}.
We compress $r_g^{(k)}$ using the proposed \textbf{Generation Residual Codec}, whose architecture is detailed in Appendix~\ref{app:generation_residual_codec}.

The codec incorporates two complementary conditioning signals.
First, we embed the prefix depth $k$ as $e_k$ and obtain a prefix-dependent residual scale $s_k$ and gain $g_k$.
These prefix-dependent parameters adapt the codec to different bitrates, since the statistics of $r_g^{(k)}$ vary with the number of omitted suffix scales.
Second, we combine autoregressive decoding features with $\hat h_q^{(k)}$ to construct a multi-scale context pyramid
$C^{(k)}=\{c_0^{(k)},c_1^{(k)},c_2^{(k)},c_3^{(k)}\}$,
which is injected into the corresponding stages of the codec.
We first normalize the generation residual $r_g^{(k)}$ by the prefix-dependent scale $s_k$ and feed it into the analysis transform:

\begin{equation}
y=g_a\!\left(r_g^{(k)}\oslash s_k;\hat h_q^{(k)},C^{(k)},e_k\right),
\end{equation}
where $\oslash$ denotes element-wise division.

Following prior works~\citep{li2023dcvcdc,sheng2025dcvcb}, we employ a four-pass conditional entropy model parameterized by $\eta$.
The analysis latent elements are first modulated by the prefix-dependent gain $g_k$ and then partitioned into four sequential coding groups, where the quantization step map is predicted once from $c_3^{(k)}$ and $e_k$ and remains fixed across the four passes. Each pass predicts the conditional mean and Laplace scale using these conditions and the latent values reconstructed in previous passes.
The quantized latent is represented by an integer-valued symbol tensor $\ddot y$, whose conditional probability mass function $p_\eta$ is used to estimate the bitrate during training:

\begin{equation}
R_g^{(k)}
=
\frac{
-\log_2 p_\eta(\ddot y\mid c_3^{(k)},e_k)
}{HW}.
\label{eq:generation_residual_rate}
\end{equation}

At the decoder, the same causal context is used to reconstruct the quantized latent $\hat y$ from $\ddot y$.
The synthesis transform then maps $\hat y$ to the reconstructed generation residual:

\begin{equation}
\hat r_g^{(k)}
=
s_k\odot
g_s\!\left(\hat y;C^{(k)},e_k\right),
\label{eq:generation_residual_reconstruction}
\end{equation}
where $\odot$ denotes element-wise multiplication.

\subsection{Residual Fusion and Reconstruction}
\label{sec:residual_fusion}
The two residual branches provide complementary information missing from the autoregressively reconstructed latent $\hat h_q^{(k)}$.
Following the decomposition in~\Cref{eq:residual_decomposition}, we obtain the compensated latent as

\begin{equation}
\hat h_c^{(k)}
=
\hat h_q^{(k)}
+
\hat r_q^{(k)}
+
\hat r_g^{(k)}.
\label{eq:residual_fusion}
\end{equation}
The adapted VAE decoder $\tilde{\mathcal D}$ then maps the compensated latent to the reconstructed image $\hat x^{(k)}=\tilde{\mathcal D}\left(\hat h_c^{(k)}\right).$

The two residual branches incur different bitrate costs.
The quantization residual is synthesized entirely from information already available at the decoder and therefore requires no additional transmitted bits, whereas the generation residual is entropy encoded and explicitly transmitted.
For the image $x\in\mathbb R^{H\times W\times3}$, the total bitrate is calculated from the measured bit lengths as

\begin{equation}
R_{\mathrm{total}}^{(k)}
=
\frac{
B_{\mathrm{text}}
+
B_{\mathrm{prefix}}^{(k)}
+
B_g^{(k)}
}{HW},
\end{equation}
where $B_{\mathrm{text}}$, $B_{\mathrm{prefix}}^{(k)}$, and $B_g^{(k)}$ denote the measured bit lengths of the text condition, the transmitted token prefix, and the generation residual bitstream, respectively.
Detailed rate accounting is provided in~\Cref{app:rate_decomposition}.

\begin{figure*}[!t]
  \centering
   
  \includegraphics[width=1.0\linewidth]{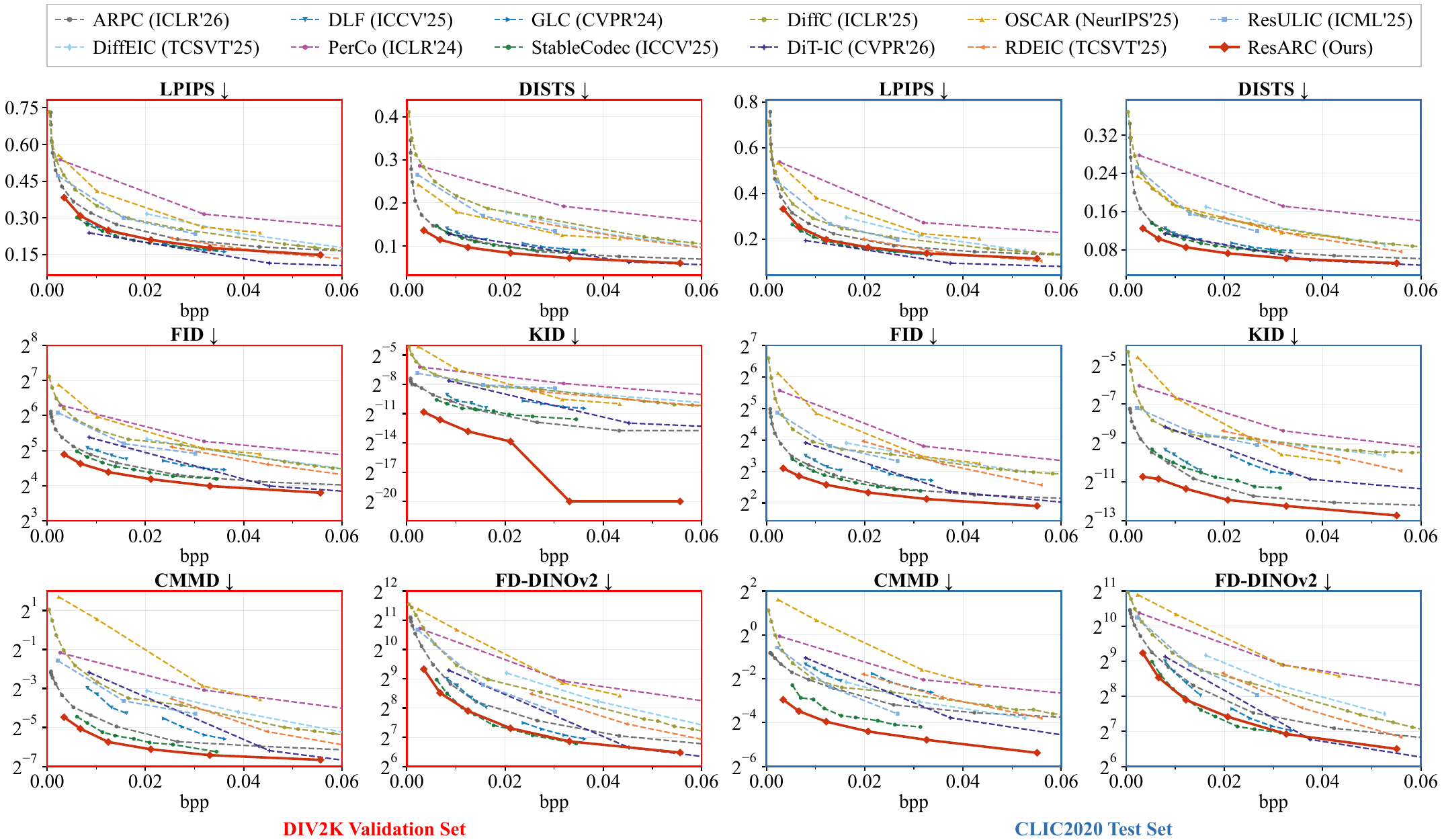}
   
  \caption{\textbf{Rate-perception comparison on DIV2K (left) and CLIC2020 (right).} 
  ResARC achieves strong overall performance across perceptual similarity and distributional fidelity metrics in the ultra-low bitrate regime.}
   
  \label{fig:resarc_all_baselines}
\end{figure*}

\subsection{Training Pipeline}
\label{sec:training_pipeline}
We train ResARC in multiple stages. Detailed training settings are provided in~\Cref{app:training_details}.
\begin{itemize}[leftmargin=1pt, itemsep=1pt, topsep=0pt, parsep=0pt]
\item \textbf{Backbone fine-tuning for compression.}
We first fine-tune the Infinity backbone~\citep{han2025infinity} on our compression training data to adapt the autoregressive model and VAE decoder from their original generative setting to the target compression setting.
To adapt the VAE decoder for quantization residual-compensated latents, we train it with the quantized latent $h_q$ and the continuous latent $h$ with equal probability, rather than only using $h_q$ as in the original backbone.
The adapted backbone is then frozen for subsequent residual learning.

\item \textbf{Quantization residual generator training.}
We next train the Quantization Residual Generator using the flow-matching objective in~\Cref{eq:quant_flow_matching}, together with perceptual supervision $\mathcal{L}_{\text{LPIPS}}$~\citep{zhang2018lpips} to improve recovery of fine-grained details.

\item \textbf{Generation residual codec training.}
We then optimize the Generation Residual Codec with a rate-distortion objective that balances the residual bitrate and reconstruction fidelity:

\begin{equation}
\mathcal{L}_{g}
=
\lambda_R R_g^{(k)}
+
\lambda_{\mathrm{res}}\mathcal{L}_{\mathrm{res}}
+
\lambda_{\mathrm{D}}\mathcal{L}_{\mathrm{DISTS}},
\end{equation}
where $\mathcal L_{\mathrm{res}}$ is the latent-space Smooth L1 loss, $\mathcal L_{\mathrm{DISTS}}$ is the image-space perceptual DISTS loss~\citep{ding2020dists}, and $\lambda_R$, $\lambda_{\mathrm{res}}$, and $\lambda_{\mathrm D}$ weight the bitrate and reconstruction terms.

\item \textbf{Residual-aware VAE decoder adaptation.}
Finally, we freeze both residual branches and adapt only the VAE decoder using the compensated latent $\hat h_c^{(k)}$.
This stage adapts the decoder to the latent distribution produced after residual compensation and improves its utilization of the recovered residual information.
\end{itemize}

\section{Experiments}

\label{sec:experiments}
\raggedbottom
\subsection{Implementation}

\paragraph{Datasets.}
We train on 1,000,000 filtered text-image pairs from COYO-700M~\citep{byeon2022coyo}.
We remove low-resolution, unsafe, watermarked, low-aesthetic, and text-heavy samples.
Autoregressive and residual learning use $1024\times1024$ crops, while the initial VAE decoder fine-tuning uses $512\times512$ crops.
For evaluation, we use the DIV2K validation set~\citep{agustsson2017div2k} with 100 images and the CLIC2020 test set~\citep{toderici2020clic} with 428 images.
Following the preprocessing protocol of ARPC~\citep{zhang2026arpc}, all evaluation images are center-cropped to $1024\times1024$.
We generate a caption for each evaluation image using Florence-2~\citep{xiao2024florence2}.

\paragraph{Training details.}
We initialize the VAE and autoregressive model from Infinity-2B~\citep{han2025infinity}, with the VAE encoder frozen throughout training.
The autoregressive transformer is fine-tuned for 2,000 iterations with a batch size of 64 and a learning rate of $6\times10^{-5}$.
We fine-tune the VAE decoder for 35,000 iterations by sampling continuous and quantized latents with equal probability.
The Quantization Residual Generator and Generation Residual Codec are then trained for 5,000 and 7,300 iterations, respectively.
Finally, with both residual branches frozen, we adapt the VAE decoder to the compensated latents for another 1,000 iterations.
All stages use AdamW~\citep{loshchilov2019decoupled}.
The training details can be found in~\Cref{app:training_details}.

\paragraph{Metrics.}
We focus on perceptual quality in the ultra-low bitrate regime from two complementary perspectives.
For paired perceptual similarity, we report LPIPS~\citep{zhang2018lpips} and DISTS~\citep{ding2020dists}, which compare each reconstruction with its corresponding original image in learned perceptual feature spaces, with DISTS further emphasizing structural and textural similarity.
For distributional fidelity, we report FID~\citep{heusel2017fid}, KID~\citep{binkowski2018kid}, CMMD~\citep{jayasumana2024cmmd}, and FD-DINOv2~\citep{stein2023exposing,oquab2023dinov2}, which measure the discrepancy between the distributions of reconstructed and original images in complementary feature spaces.
Lower values indicate better performance for all six metrics.

\paragraph{Baselines.}
We compare ResARC with eleven leading open-source generative image codecs at their released bitrate points.
These baselines include the autoregressive codec ARPC~\citep{zhang2026arpc}; diffusion-based codecs DiffEIC~\citep{li2025diffeic}, DiffC~\citep{vonderfecht2025diffc}, DiT-IC~\citep{shi2026ditic}, OSCAR~\citep{guo2025oscar}, PerCo~\citep{careil2024perco}, RDEIC~\citep{li2025rdeic}, ResULIC~\citep{ke2025resulic}, and StableCodec~\citep{zhang2025stablecodec}; and generative latent codecs DLF~\citep{xue2025dlf} and GLC~\citep{jia2024glc}.
For all methods, the reported bitrate includes all transmitted components specified by the corresponding codec.

\begin{figure*}[!t]
    \centering
    \includegraphics[width=1.0\linewidth]{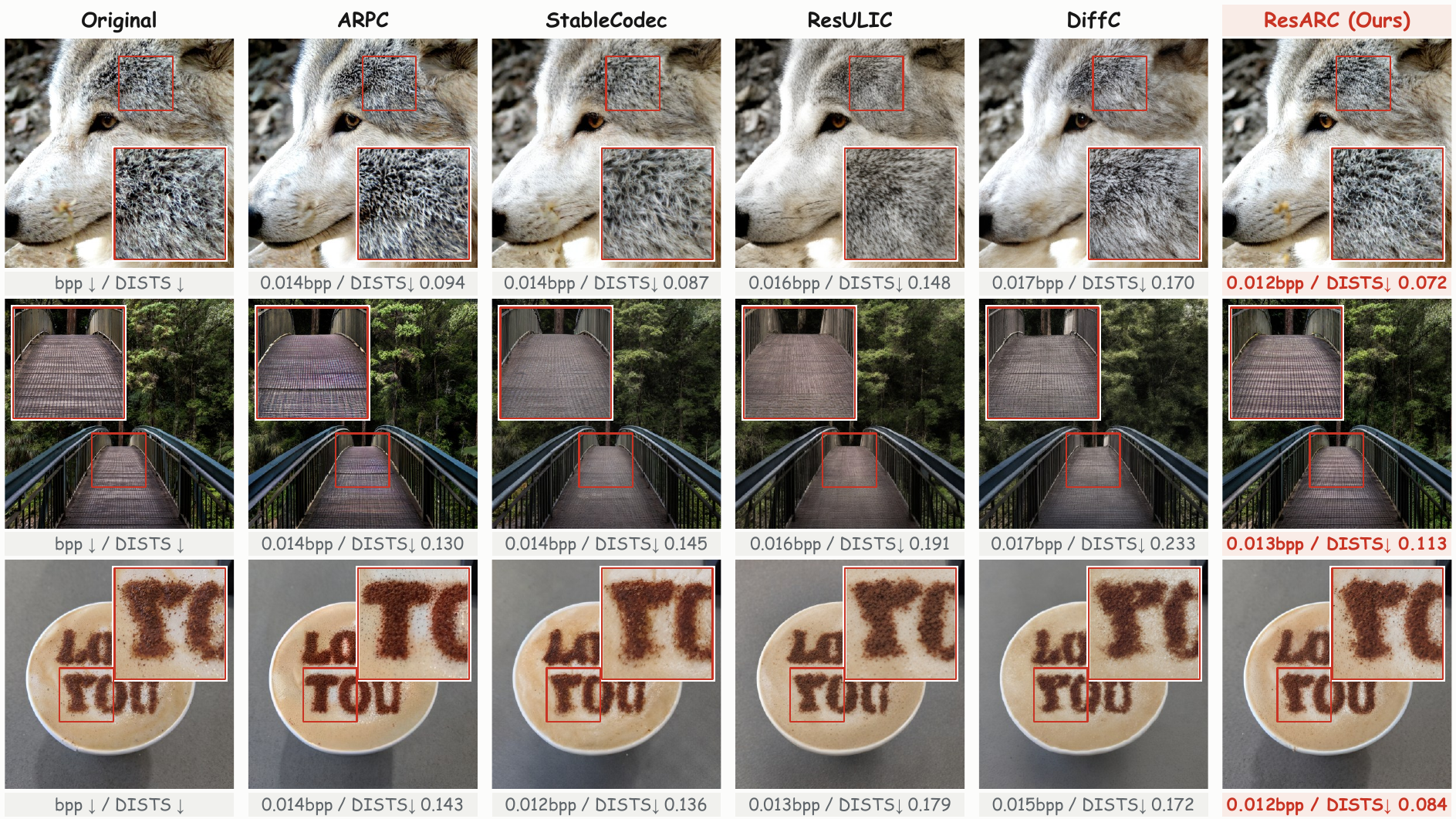}
    \caption{\textbf{Qualitative comparison with other generative codecs on DIV2K and CLIC2020.}}
    \label{fig:resarc_visual_comparison}
\end{figure*}

\subsection{Quantitative Comparisons}

\Cref{fig:resarc_all_baselines} compares ResARC with leading generative codecs on the DIV2K validation set and CLIC2020 test set.
Across both datasets, ResARC achieves a favorable rate-perception trade-off throughout the ultra-low bitrate regime.
Compared with the autoregressive codec ARPC, ResARC consistently improves all six evaluated metrics over the evaluated bitrate range.
Across all compared methods, ResARC achieves the strongest overall performance under DISTS and the distributional fidelity metrics FID, KID, CMMD, and FD-DINOv2, while remaining competitive under LPIPS.
These results demonstrate that compensating for both quantization and generation residuals substantially improves the reconstruction fidelity of autoregressive generative compression.
The complete signed BD-rate comparison is provided in Appendix~\ref{app:bd_rate_comparison}, further validating the consistent improvements of ResARC across metrics and datasets.

\subsection{Qualitative Comparisons}

\Cref{fig:resarc_visual_comparison} compares ResARC with leading generative codecs at similar ultra-low bitrates.
The autoregressive codec ARPC~\citep{zhang2026arpc} produces rich textures but may alter image structures, such as fur patterns, repeated lines on the bridge deck, and lettering on the coffee surface, likely due to imperfect suffix generation.
In contrast, diffusion-based methods such as StableCodec~\citep{zhang2025stablecodec} and ResULIC~\citep{ke2025resulic} tend to smooth these local details, reducing their similarity to the original images.
ResARC produces less over-smoothed reconstructions and better preserves fine details and image structures, including the arrangement of wolf fur, repeated bridge-deck patterns, and the shape of the lettering.
These qualitative results are consistent with our residual-aware design, where quantization residual generation helps recover fine-grained details, while generation residual correction helps preserve image structures and textures.
Additional qualitative comparisons are provided in~\Cref{app:additional_visuals}.

\subsection{Ablation Studies}
\label{sec:ablation_studies}
We ablate the contributions of the two residuals, their compensation strategies and training objectives, and the training pipeline, including fine-tuning the autoregressive prior, residual-aware VAE decoder adaptation, and the decoder input distribution used for fine-tuning. 
All experiments are evaluated on the DIV2K validation set. 
Tables report signed BD-rate (\%) relative to the specified reference, where lower values are better.

\setcounter{topnumber}{1}
\setcounter{totalnumber}{1}
\renewcommand{\topfraction}{0.75}
\renewcommand{\textfraction}{0.25}

\begin{figure}[t]
\centering
\begin{minipage}{1.0\linewidth}
\centering
\small

\setlength{\abovecaptionskip}{2pt}
\setlength{\belowcaptionskip}{2pt}

\captionof{table}{\textbf{Ablation study on the contribution of the quantization and generation residuals.}}
\label{tab:residual_component_bd}

\setlength{\tabcolsep}{4pt}
\renewcommand{\arraystretch}{1.08}

\begin{tabularx}{0.95\linewidth}{l*{6}{>{\centering\arraybackslash}X}}
\toprule
Configuration & $r_q$ & $r_g$
& LPIPS$\downarrow$ & DISTS$\downarrow$
& FID$\downarrow$ & KID$\downarrow$ \\
\midrule

Autoregressive Baseline
& \xmark & \xmark & +7.78 & +9.80 & +26.57 & +340.22 \\

w/ Generation Residual Branch
& \xmark & \cmark & +2.42 & +1.90 & +8.03 & +67.91 \\

w/ Quantization Residual Branch
& \cmark & \xmark & +3.46 & +5.38 & +23.99 & +263.50 \\

\rowcolor{rowblue}
\textbf{ResARC (Both Residual Branches)}
& \cmark & \cmark
& \textbf{0.00} & \textbf{0.00}
& \textbf{0.00} & \textbf{0.00} \\

\bottomrule
\end{tabularx}
\end{minipage}
\par\medskip
\centering
\begin{minipage}[t]{0.40\textwidth}
\centering
\includegraphics[width=0.95\linewidth]
{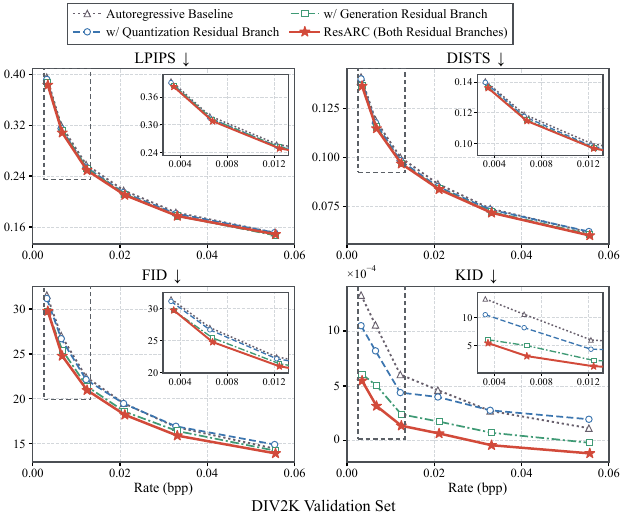}
\captionof{figure}{\protect\raggedright \textbf{Rate-perception curves for the contribution of the quantization and generation residuals on DIV2K.}}
\label{fig:residual_component_rd_div2k}
\end{minipage}
\hfill
\begin{minipage}[t]{0.59\textwidth}
\centering
\includegraphics[pagebox=cropbox,clip,width=\linewidth]
{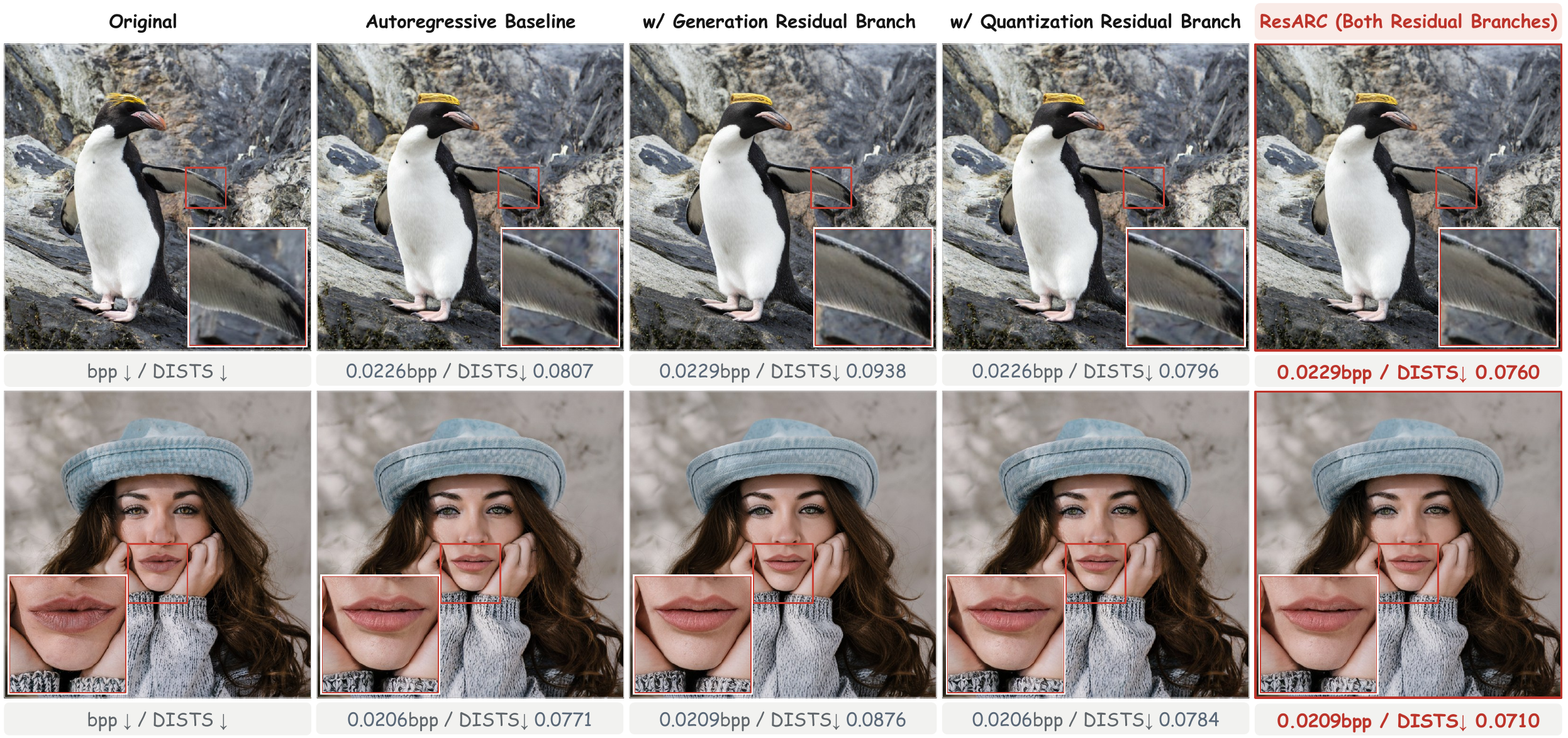}
\vspace{-0.4cm}
\captionof{figure}{\protect\raggedright \textbf{Qualitative illustration of the complementary contributions of the two residual branches.}}
\label{fig:residual_component_visual}
\end{minipage}
\end{figure}

\paragraph{Contribution of the two residuals.}
Starting from the autoregressive baseline, we separately add the generation residual branch and the quantization residual branch, and then combine both in full ResARC.
As shown in~\Cref{tab:residual_component_bd}, either residual branch improves rate-perception performance over the autoregressive baseline, while combining both yields the best overall performance.
Specifically, with only the generation residual branch, the DISTS and FID BD-rates decrease from $+9.80\%$ and $+26.57\%$ to $+1.90\%$ and $+8.03\%$, respectively; with only the quantization residual branch, they decrease to $+5.38\%$ and $+23.99\%$.
Full ResARC further improves all four metrics, demonstrating the complementary benefits of the two residual branches.

We further provide results on the contribution of the two residual branches through rate-perception curves.
For all configurations, we follow the same training stages and optimization settings, while only adding the corresponding residual branch to the autoregressive baseline.
As shown in~\Cref{fig:residual_component_rd_div2k}, adding either the generation residual branch or the quantization residual branch consistently improves performance across all four metrics.
Moreover, combining both residual branches in full ResARC yields further improvements and achieves the best overall performance, demonstrating the complementary benefits of the two residuals.

\Cref{fig:residual_component_visual} further illustrates the complementary roles of the two residual branches.
The enlarged regions show differences in the reconstructed feather details of the penguin and the mouth contours of the portrait.
Combining both branches yields reconstructions that better preserve local details and image structures, and achieves the lowest DISTS in the displayed examples.

\begin{figure}[t]
\centering
\begingroup
\setlength{\abovecaptionskip}{3.5pt}
\setlength{\belowcaptionskip}{3.5pt}

\begin{minipage}[t]{0.495\textwidth}
\centering
\footnotesize

\captionof{table}{\textbf{Effect of fine-tuning the AR prior.}}
\label{tab:infinity_ft_bd}

\setlength{\tabcolsep}{0.8pt}
\renewcommand{\arraystretch}{1.08}
\begin{tabularx}{\linewidth}{l*{4}{>{\centering\arraybackslash}X}}
\toprule
Configuration & LPIPS$\downarrow$ & DISTS$\downarrow$ & FID$\downarrow$ & KID$\downarrow$ \\
\midrule
w/o AR fine-tuning & 0.00 & 0.00 & 0.00 & 0.00 \\
\rowcolor{rowblue}
\textbf{w/ AR fine-tuning}
& \textbf{-4.56}
& \textbf{-6.42}
& \textbf{-11.29}
& \textbf{-29.67} \\
\bottomrule
\end{tabularx}
\end{minipage}
\hfill
\begin{minipage}[t]{0.495\textwidth}
\centering
\footnotesize

\captionof{table}{\textbf{Effect of VAE decoder adaptation.}}
\label{tab:decoder_adaptation_div2k}

\setlength{\tabcolsep}{0.3pt}
\renewcommand{\arraystretch}{1.08}
\begin{tabularx}{\linewidth}{l*{4}{>{\centering\arraybackslash}X}}
\toprule
Configuration & LPIPS$\downarrow$ & DISTS$\downarrow$ & FID$\downarrow$ & KID$\downarrow$ \\
\midrule
w/o Adapt VAE Dec. & 0.00 & 0.00 & 0.00 & 0.00 \\
\rowcolor{rowblue}
\textbf{w/ Adapt VAE Dec.}
& \textbf{-4.31}
& \textbf{-7.19}
& \textbf{-4.24}
& \textbf{-24.20} \\
\bottomrule
\end{tabularx}
\end{minipage}

\par\medskip

\begin{minipage}{\linewidth}
\centering
\small

\captionof{table}{\textbf{Ablation of residual compensation strategies.}
BD-rate (\%) relative to ResARC (Ours).}
\label{tab:operator_assignment_decoder1k}

\setlength{\tabcolsep}{3pt}
\renewcommand{\arraystretch}{1.08}
\begin{tabularx}{0.98\linewidth}{l*{4}{>{\centering\arraybackslash}X}}
\toprule
Configuration & LPIPS$\downarrow$ & DISTS$\downarrow$ & FID$\downarrow$ & KID$\downarrow$ \\
\midrule
Generate both residuals
& \textbf{-0.75} & +10.46 & +11.11 & +27.34 \\
Transmit both residuals
& +4.67 & +6.29 & +6.99 & +35.65 \\
\rowcolor{rowblue}
\textbf{Generate $r_q$ + transmit $r_g$ (Ours)}
& 0.00 & \textbf{0.00} & \textbf{0.00} & \textbf{0.00} \\
\bottomrule
\end{tabularx}
\end{minipage}

\par\medskip

\begin{minipage}[t]{0.44\textwidth}
\centering
\includegraphics[width=0.955\linewidth]
{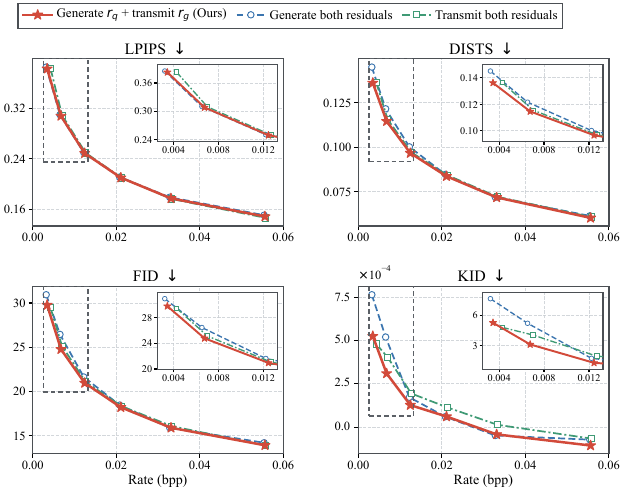}
\captionof{figure}{\textbf{Rate-perception curves for residual compensation strategies.}}
\label{fig:operator_assignment_decoder1k}
\end{minipage}
\hfill
\begin{minipage}[t]{0.55\textwidth}
\centering
\includegraphics[pagebox=cropbox,clip,width=\linewidth]
{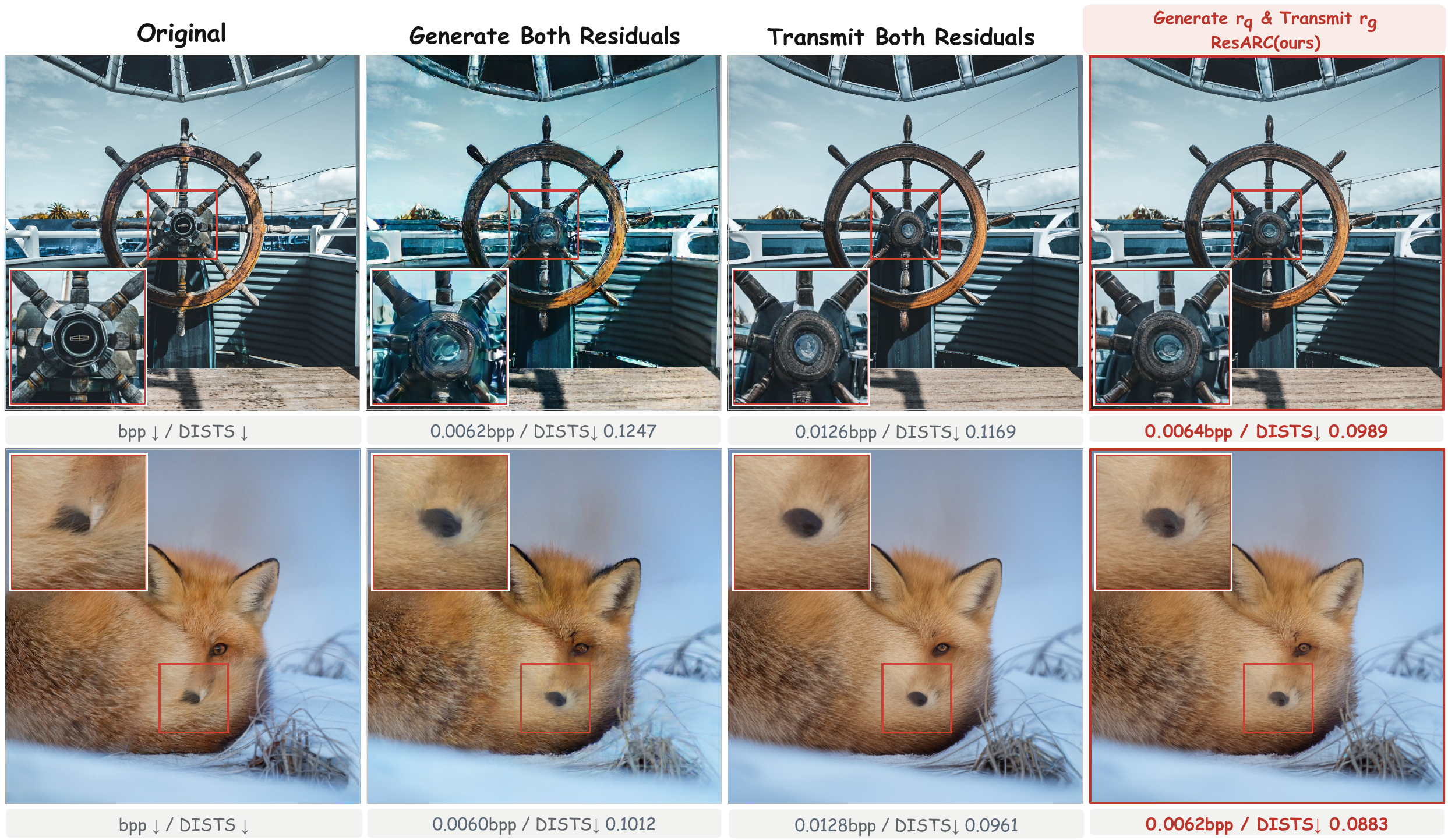}
\vspace{-0.4cm}
\captionof{figure}{\textbf{Visual comparison of residual compensation strategies.}}
\label{fig:operator_assignment_visual}
\end{minipage}

\endgroup
\end{figure}

\paragraph{Fine-tuning the autoregressive prior.}
To isolate the effect of adapting the autoregressive prior to the compression task, we compare the pretrained Infinity-2B transformer~\citep{han2025infinity} with our model fine-tuned on the compression training data.
As shown in~\Cref{tab:infinity_ft_bd}, fine-tuning consistently improves all four metrics on DIV2K, yielding BD-rate reductions of $6.42\%$ under DISTS and $11.29\%$ under FID.
These results show that adapting the autoregressive prior provides a stronger foundation for subsequent training of the residual branches.

\paragraph{Residual-aware VAE decoder adaptation.}
We evaluate the effect of adapting the VAE decoder to compensated latents by comparing perceptual quality before and after adaptation.
As shown in~\Cref{tab:decoder_adaptation_div2k}, the residual-aware adaptation improves all four metrics on DIV2K, including BD-rate reductions of $4.31\%$ under LPIPS and $7.19\%$ under DISTS.
These results demonstrate that adapting the VAE decoder to the compensated latent improves perceptual reconstruction quality.

\paragraph{Residual compensation strategies.}
We compare our residual-specific strategy, which generates the quantization residual $r_q$ with a conditional DiT and transmits the generation residual $r_g$ with a learned codec, against two uniform strategies that either generate or transmit both residuals.
All experiments use the same two residual targets, autoregressive backbone, and sampling settings.
As shown in~\Cref{tab:operator_assignment_decoder1k}, relative to ResARC, generating both residuals incurs $10.46\%$ and $11.11\%$ higher BD-rates under DISTS and FID, respectively, despite a slight improvement under LPIPS, while transmitting both residuals increases BD-rate across all four metrics.
These results support the residual-specific compensation design of ResARC, where the quantization residual $r_q$ is generated from decoder-side context without additional transmitted bits, while $r_g$ is explicitly transmitted to provide source-dependent correction.

Beyond the aggregate BD-rate results, we further compare the complete rate--perception curves in~\Cref{fig:operator_assignment_decoder1k}. 
Across the evaluated bitrate range, transmitting both residuals consistently underperforms ResARC on all four metrics, while generating both achieves slightly better LPIPS but worse DISTS, FID, and KID. 
These  results further support the residual-specific design of ResARC: the quantization residual can be effectively inferred from decoder-side autoregressive context without additional transmitted bits, whereas the generation residual depends on source suffix information unavailable at the decoder and therefore benefits from explicit transmission.

We further provide a qualitative comparison of the three strategies in~\Cref{fig:operator_assignment_visual}.
ResARC better preserves local structures and fine details, such as the fox's nose and the central hub of the ship's wheel, further supporting the effectiveness of our residual-specific compensation design.

\begin{figure}[t]
\centering
\begin{minipage}{\linewidth}
\centering
\small
\captionof{table}{\protect\raggedright
\textbf{Training-objective ablations on DIV2K.}
Signed BD-rate (\%) is reported relative to the objective used in ResARC (Ours) within each branch. FM denotes flow matching; - denotes no common quality interval.}
\label{tab:branch_loss_bd_div2k}
\setlength{\tabcolsep}{5pt}
\renewcommand{\arraystretch}{1.09}
\begin{tabularx}{0.9\linewidth}{Xcccc}
\toprule
Training objective & LPIPS$\downarrow$ & DISTS$\downarrow$ & FID$\downarrow$ & KID$\downarrow$ \\
\midrule
\multicolumn{5}{l}{\textit{Quantization residual generator}} \\
\arrayrulecolor{black!30}\midrule
FM
  & $+3.38$ & $+2.60$ & $+7.62$ & $+45.37$ \\
FM + DISTS
  & $+2.81$ & $-0.25$ & $+4.33$ & $+32.50$ \\
FM + LPIPS + DISTS
  & $+1.70$ & $\mathbf{-0.65}$ & $+1.27$ & $+12.10$ \\
\rowcolor{rowblue}
\textbf{FM + LPIPS (Ours)}
  & $\mathbf{0.00}$ & $0.00$ & $\mathbf{0.00}$ & $\mathbf{0.00}$ \\
\addlinespace[2pt]
\arrayrulecolor{black}\midrule
\multicolumn{5}{l}{\textit{Generation residual codec}} \\
\arrayrulecolor{black!30}\midrule
Rate + Smooth L1
  & $+64.27$ & $+155.82$ & $+278.69$ & - \\
Rate + Smooth L1 + LPIPS
  & $+62.97$ & $+194.08$ & $+233.80$ & $+787.71$ \\
Rate + Smooth L1 + LPIPS + DISTS
  & $\mathbf{-6.40}$ & $+9.69$ & $+19.50$ & $+79.17$ \\
\rowcolor{rowblue}
\textbf{Rate + Smooth L1 + DISTS (Ours)}
  & $0.00$ & $\mathbf{0.00}$ & $\mathbf{0.00}$ & $\mathbf{0.00}$ \\
\arrayrulecolor{black}\bottomrule
\end{tabularx}
\end{minipage}
\par\medskip
\begin{minipage}{\linewidth}
\centering
\small
\captionof{table}{\protect\raggedright \textbf{Ablation of decoder input distributions during fine-tuning.}
BD-rate (\%) is reported relative to mixed fine-tuning (\(p=0.5\)).}
\label{tab:decoder_domain_both}
\setlength{\tabcolsep}{10pt}
\begin{tabular}{lcccc}
\toprule
Type
& LPIPS$\downarrow$ & DISTS$\downarrow$
& FID$\downarrow$ & KID$\downarrow$ \\
\midrule
Quantized ($p=0$)
& +0.06
& +7.42
& +4.03
& +0.73 \\
Continuous ($p=1$)
& +4.45 & +31.99 & +52.80 & +204.28 \\
\rowcolor{rowblue}
Mixed ($p=0.5$)
& \textbf{0.00}
& \textbf{0.00}
& \textbf{0.00}
& \textbf{0.00} \\
\bottomrule
\end{tabular}
\end{minipage}
\end{figure}

\paragraph{Training Objectives for the Residual Branches.}
\label{app:branch_objectives}

We study the perceptual objectives used to train the two residual branches.
For each branch, we vary only the perceptual loss while keeping the remaining training objective fixed: flow matching for the quantization residual generator, and rate together with Smooth L1 loss for the generation residual codec.

\Cref{tab:branch_loss_bd_div2k} reports BD-rates relative to the objective used in ResARC for each branch.
For the Quantization Residual Generator, using the LPIPS loss yields the best LPIPS, FID, and KID BD-rates among the tested objectives.
However, adding DISTS supervision improves the DISTS BD-rate by only $0.65\%$, while increasing the LPIPS, FID, and KID BD-rates by $1.70\%$, $1.27\%$, and $12.10\%$, respectively.
For the Generation Residual Codec, adding the LPIPS loss to the DISTS objective reduces the LPIPS BD-rate by $6.40\%$, but increases the DISTS, FID, and KID BD-rates by $9.69\%$, $19.50\%$, and $79.17\%$, respectively.
These results support using the LPIPS objective for the quantization residual generator and DISTS loss for the generation residual codec.

\paragraph{Mixed-Latent VAE Decoder Fine-Tuning.}
\label{app:decoder_preparation}
We further investigate the input distribution used to fine-tune the VAE decoder.
Quantization residual compensation shifts the decoder input from the quantized latent $h_q$ toward the continuous latent $h$.
We therefore compare three fine-tuning settings: quantized latents only, continuous latents only, and a mixed setting with $p=0.5$, where $p$ denotes the probability of sampling $h$.
As shown in~\Cref{tab:decoder_domain_both}, the mixed setting achieves the best overall BD-rate across all four metrics.
Compared with mixed-latent fine-tuning, using only quantized latents incurs $7.42\%$ and $4.03\%$ higher BD-rates under DISTS and FID, respectively, while using only continuous latents leads to substantially higher BD-rates across all four metrics.
These results support mixed-latent fine-tuning for adapting the VAE decoder to compensated latent representations.

\section{Conclusion}
We identify two residuals inherent to autoregressive generative compression: the \emph{quantization residual} introduced by discrete tokenization and the \emph{generation residual} arising from imperfect autoregressive suffix generation.
To recover both residuals, we introduce \textbf{ResARC}, the first residual-aware autoregressive codec that generates the quantization residual with a Quantization Residual Generator based on DiT and compresses the generation residual with a dedicated Generation Residual Codec for transmission.
Extensive experiments demonstrate the effectiveness of this residual-aware design, with ResARC achieving strong rate-perception performance compared with leading generative codecs.
These results highlight explicit modeling and compensation of codec residuals as an effective design principle for ultra-low bitrate generative image compression.
\par
\endgroup

\bibliographystyle{iclr2027_conference}
\bibliography{iclr2027_conference}

\clearpage
\appendix
\begingroup

\setlength{\textfloatsep}{9pt plus 2pt minus 2pt}
\setlength{\intextsep}{8pt plus 2pt minus 2pt}
\setlength{\floatsep}{8pt plus 2pt minus 2pt}
\setlength{\abovecaptionskip}{4pt}
\setlength{\belowcaptionskip}{3pt}

\section{Pseudo-Code of ResARC}
\label{app:codec_algorithm}

\Cref{alg:resarc_codec} and \Cref{alg:resarc_quant_sampler} summarize the ResARC encoding-decoding pipeline and quantization residual generation, respectively.

\subsection{Encoding and Decoding}
\label{app:encoding_decoding}
During encoding, the encoder reproduces the same autoregressive suffix generation process used at the decoder from the transmitted prefix $T_{\leq k}$ and text condition $\ell$.
This reconstructs $\hat h_q^{(k)}$, from which the generation residual
$r_g^{(k)}=h_q-\hat h_q^{(k)}$ is computed and compressed by the Generation Residual Codec.
The resulting bitstream contains the encoded text condition, transmitted prefix scales, and entropy-coded generation residual.

During decoding, the text condition and prefix scales are first recovered and used to regenerate the same suffix and reconstruct $\hat h_q^{(k)}$.
The autoregressive decoding process also provides the condition $c_{\mathrm{quant}}$ for synthesizing the quantization residual and context $C^{(k)}$ for the generation residual.
To ensure identical autoregressive states at the encoder and decoder, \textsc{GenerateSuffix} uses the same sampling realization at both endpoints, while residual entropy coding follows the same causal symbol order.
Finally, the decoder generates the quantization residual $\hat r_q^{(k)}$, decodes the transmitted generation residual $\hat r_g^{(k)}$, and combines both with $\hat h_q^{(k)}$ to form the compensated latent $\hat h_c^{(k)}$.
The adapted VAE decoder $\tilde{\mathcal D}$ then maps $\hat h_c^{(k)}$ to the reconstructed image.

\begin{algorithm}[H]
\caption{ResARC encoding and decoding}
\label{alg:resarc_codec}
\small
\begin{algorithmic}[1]
\Require Shared models and fixed sampling settings in encoding and decoding
\Procedure{Encode}{$x,\ell,k$}
    \State $h\gets\mathcal{E}(x)$; $\mathcal{T}\gets\mathcal{Q}(h)$; $h_q\gets\mathcal{S}(\mathcal{T})$
    \State $\mathcal{B}_{\mathrm{text}}\gets\Call{EncodeText}{\ell}$
    \State $\mathcal{B}_{\mathrm{prefix}}\gets\Call{EncodePrefix}{T_{\leq k},\ell;p_\psi}$
    \State $\hat T_{>k}\gets\Call{GenerateSuffix}{T_{\leq k},\ell;p_\psi}$
    \State $\hat h_q^{(k)}\gets\mathcal{S}(T_{\leq k},\hat T_{>k})$
    \State Construct $C^{(k)}$ from the autoregressive model
    \State Construct $e_k,s_k,g_k$ from $k$
    \State $r_g^{(k)}\gets h_q-\hat h_q^{(k)}$
    \State $y\gets g_a(r_g^{(k)}\oslash s_k;\hat h_q^{(k)},C^{(k)},e_k)$
    \State $\ddot y\gets\mathcal Q_\eta(g_k\odot y\mid c_3^{(k)},e_k)$
    \State $\mathcal B_g\gets\Call{EntropyEncode}{\ddot y;p_\eta(\ddot y\mid c_3^{(k)},e_k)}$
    \State \Return $\mathcal{B}\gets\Call{Assemble}{k,\mathcal{B}_{\mathrm{text}},\mathcal{B}_{\mathrm{prefix}},\mathcal{B}_g}$
\EndProcedure
\Statex
\Procedure{Decode}{$\mathcal{B}$}
    \State $(k,\mathcal{B}_{\mathrm{text}},\mathcal{B}_{\mathrm{prefix}},\mathcal{B}_g)\gets\Call{Parse}{\mathcal{B}}$
    \State $\ell\gets\Call{DecodeText}{\mathcal{B}_{\mathrm{text}}}$
    \State $T_{\leq k}\gets\Call{DecodePrefix}{\mathcal{B}_{\mathrm{prefix}},\ell;p_\psi}$
    \State $\hat T_{>k}\gets\Call{GenerateSuffix}{T_{\leq k},\ell;p_\psi}$
    \State $\hat h_q^{(k)}\gets\mathcal{S}(T_{\leq k},\hat T_{>k})$
    \State Construct $c_{\mathrm{quant}}$ and $C^{(k)}$ from the autoregressive model
    \State Construct $e_k,s_k,g_k$ from $k$
    \State $\ddot y\gets\Call{EntropyDecode}{\mathcal B_g;p_\eta(\ddot y\mid c_3^{(k)},e_k)}$
    \State $\hat u\gets\mathcal R_\eta(\ddot y\mid c_3^{(k)},e_k)$
    \State $\hat y\gets\hat u\oslash g_k$
    \State $\hat r_g^{(k)}\gets s_k\odot g_s(\hat y;C^{(k)},e_k)$
    \State Draw $\epsilon\sim\mathcal{N}(0,I)$
    \State $\hat r_q^{(k)}\gets\Call{GenerateQuantizationResidual}{c_\mathrm{quant},\epsilon,N}$
    \State $\hat{h}_c^{(k)}\gets\hat h_q^{(k)}+\hat r_q^{(k)}+\hat r_g^{(k)}$
    \State \Return $\hat x^{(k)}\gets\tilde{\mathcal{D}}(\hat{h}_c^{(k)})$
\EndProcedure
\end{algorithmic}
\end{algorithm}

\begin{algorithm}[t]
\caption{Quantization residual generation}
\label{alg:resarc_quant_sampler}
\small
\begin{algorithmic}[1]
\Require Diffusion Transformer $v_\theta$
\Procedure{GenerateQuantizationResidual}{$c_{\mathrm{quant}},\epsilon,N$}
    \State $z\gets\epsilon$
    \For{$j=0,\ldots,N-1$}
        \State $t_j\gets1-j/N$; $t_{j+1}\gets1-(j+1)/N$
        \State $\Delta t\gets t_{j+1}-t_j$
        \State $v\gets v_\theta(z,t_j;c_{\mathrm{quant}})$
        \State $\tilde z\gets z+\Delta t\,v$
        \If{$j<N-1$}
            \State $\tilde v\gets v_\theta(\tilde z,t_{j+1};c_{\mathrm{quant}})$
            \State $z\gets z+\frac{\Delta t}{2}(v+\tilde v)$
        \Else
            \State $z\gets\tilde z$
        \EndIf
    \EndFor
    \State \Return $\hat r_q^{(k)}\gets z$
\EndProcedure
\end{algorithmic}
\end{algorithm}

\begin{figure}[t]
  \centering
  \includegraphics[width=1.0\linewidth]{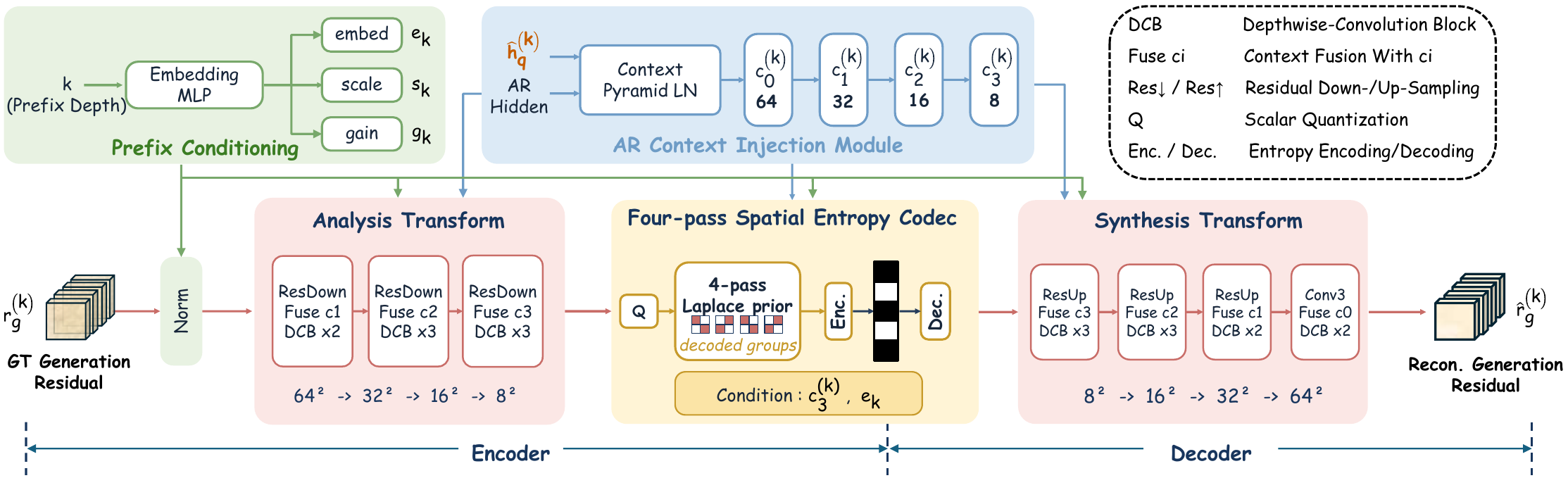}
  
  \caption{\textbf{Architecture of the Generation Residual Codec.}
  Prefix-depth conditioning and autoregressive context guide the analysis, entropy modeling, and synthesis stages of generation residual coding.}
  \label{fig:resarc_generation_residual_codec}
\end{figure}

\subsection{Quantization Residual Generation}
\label{app:shared_state}

\Cref{alg:resarc_quant_sampler} summarizes the sampling procedure of the Quantization Residual Generator. 
Starting from Gaussian noise, we integrate the conditional velocity field from $t=1$ to $t=0$ using Heun updates with an Euler update on the final step. 
At each step, the DiT predicts the velocity conditioned on the autoregressive decoding context $c_{\mathrm{quant}}$, and the final sample $z_0$ is the predicted quantization residual $\hat r_q^{(k)}$.

\section{Generation Residual Codec}
\label{app:generation_residual_codec}

\paragraph{Architecture.}
\Cref{fig:resarc_generation_residual_codec} illustrates the Generation Residual Codec introduced in~\Cref{sec:generation_residual_coding}.
The statistics of the generation residual $r_g^{(k)}$ vary with the transmitted prefix depth $k$, since different numbers of suffix scales are generated autoregressively.
To adapt the codec across different prefix depths, we introduce a prefix-depth embedding $e_k$, together with a residual scale $s_k$ and latent gain $g_k$.
The embedding $e_k$ conditions the analysis, entropy, and synthesis transforms, while $s_k$ normalizes the generation residual and $g_k$ modulates the codec latent before quantization.
In addition, the autoregressively reconstructed latent $\hat h_q^{(k)}$ and the hidden features of the autoregressive model provide context for modeling $r_g^{(k)}$.
We aggregate these features into a multi-scale context pyramid $C^{(k)}$, which is injected into the corresponding stages of the codec.
Conditioned on $e_k$ and $C^{(k)}$, the Generation Residual Codec consists of an analysis transform, a four-pass conditional entropy model, and a synthesis transform for compressing $r_g^{(k)}$ and reconstructing $\hat r_g^{(k)}$.

\paragraph{Four-pass entropy model.}
Following prior works~\citep{li2023dcvcdc,sheng2025dcvcb}, we adopt a four-pass channel-spatial entropy model for the analysis latent $y$.
The channels of $y$ are first divided into four equal groups.
Within each channel group, spatial positions are further partitioned according to the row-column parity masks $M_0,M_1,M_2,M_3$, corresponding to $(0,0)$, $(0,1)$, $(1,0)$, and $(1,1)$ under zero-based coordinates.
The four groups follow different mask orders across the four passes:
\begin{center}
\small
\setlength{\tabcolsep}{5pt}
\renewcommand{\arraystretch}{1.12}
\begin{tabularx}{0.78\linewidth}{c*{4}{>{\centering\arraybackslash}X}}
\toprule
\rowcolor{rowblue}
\textbf{Pass} & \textbf{Group 0} & \textbf{Group 1}
             & \textbf{Group 2} & \textbf{Group 3} \\
\midrule
1 & $M_0$ & $M_1$ & $M_2$ & $M_3$ \\
2 & $M_3$ & $M_2$ & $M_1$ & $M_0$ \\
3 & $M_2$ & $M_3$ & $M_0$ & $M_1$ \\
4 & $M_1$ & $M_0$ & $M_3$ & $M_2$ \\
\bottomrule
\end{tabularx}
\end{center}
This schedule assigns each latent element to exactly one coding pass while allowing all elements within the same pass to be processed in parallel.

Before quantization, the analysis latent $y$ is modulated by the prefix-dependent gain $g_k$.
A conditional prior predicts a positive quantization-step map $q$, together with the initial mean $\mu$ and Laplace scale $\sigma$, from the autoregressive context $c_3^{(k)}$ and prefix embedding $e_k$.
The quantization-step map $q$ remains fixed across the four passes.
For later passes, the mean and scale parameters $\mu,\sigma$ are further updated using the latent values reconstructed in previous passes, while positions that have not yet been decoded are masked to zero.

These predicted parameters determine the quantization and reconstruction of each latent element.
Let $\mathcal I_j$ denote the index set of elements in $y$ assigned to coding pass $j$.
For $i\in\mathcal I_j$, the integer symbol $\ddot y_i$ and corresponding reconstructed latent value $\hat y_i$ are given by
\begin{equation}
\ddot y_i
=
Q\!\left(
\frac{g_{k,i}y_i}{q_i}
-
\mu_{j,i}
\right),
\qquad
\hat y_i
=
\frac{q_i}{g_{k,i}}
\left(
\ddot y_i+\mu_{j,i}
\right),
\end{equation}
where $Q(v)=\operatorname{round}(v)$ denotes the quantizer, $g_{k,i}$ denotes the $i$-th element of the prefix-dependent gain $g_k$, and $\mu_{j,i}$ denotes the conditional mean predicted for pass $j$.
The resulting integer-valued tensor $\ddot y$ is entropy-coded for transmission. Causal context updates use the reconstructed normalized values $\ddot y_i+\mu_{j,i}$, whereas the rescaled latent $\hat y$ is passed to the synthesis transform.

For entropy modeling, each symbol is modeled with a discretized zero-mean Laplace distribution centered by the predicted conditional mean $\mu$.
For element $i$ coded in pass $j$, the probability mass assigned to integer symbol $n$ is
\begin{equation}
P_{j,i}(n)
=
F_{\sigma_{j,i}}\!\left(n+\tfrac{1}{2}\right)
-
F_{\sigma_{j,i}}\!\left(n-\tfrac{1}{2}\right),
\end{equation}
where $F_{\sigma}$ denotes the CDF of a zero-mean Laplace distribution with predicted scale $\sigma_{j,i}$.

The conditional probability of the complete symbol tensor $\ddot y$ is then factorized according to the four-pass causal coding order:
\begin{equation}
p_\eta\!\left(\ddot y\mid c_3^{(k)},e_k\right)
=
\prod_{j=1}^{4}
\prod_{i\in\mathcal I_j}
P_{j,i}\!\left(\ddot y_i\right),
\end{equation}
where the parameters of $P_{j,i}$ for each pass are conditioned on $c_3^{(k)}$, $e_k$, and the symbols reconstructed in previous passes.
Accordingly, the generation-residual bitrate during training is estimated as
\begin{equation}
R_g^{(k)}
=
\frac{
-\log_2 p_\eta\!\left(\ddot y\mid c_3^{(k)},e_k\right)
}{HW}.
\label{eq:appendix_generation_residual_rate}
\end{equation}

At the decoder, the same pass order and causal context updates are used to recover $\hat y$ from the decoded symbols $\ddot y$.
The synthesis transform then reconstructs the generation residual as
\begin{equation}
\hat r_g^{(k)}
=
s_k\odot
g_s\!\left(\hat y;C^{(k)},e_k\right).
\label{eq:appendix_generation_residual_reconstruction}
\end{equation}

\paragraph{Training overview.}
Beyond latent-space residual reconstruction, we also optimize the perceptual quality of the reconstructed image.
Specifically, we combine $\hat r_g^{(k)}$ with the autoregressively reconstructed latent $\hat h_q^{(k)}$ and decode
\begin{equation}
\hat x_g^{(k)}
=
\mathcal D_0\!\left(\hat h_q^{(k)}+\hat r_g^{(k)}\right),
\end{equation}
which is used for image-space perceptual supervision. Here, $\mathcal D_0$ denotes the VAE decoder obtained by mixed-latent fine-tuning and kept frozen during residual-branch training.
Based on both latent-space and image-space supervision, we train the codec in three stages: latent-space rate-distortion training, image-space perceptual refinement, and joint fine-tuning.
Training details for the Generation Residual Codec are provided in~\Cref{app:training_details}.

\section{Training Details}
\label{app:training_details}
\label{app:staged_optimization}
We train ResARC in stages: first fine-tuning the backbone, then optimizing the two residual branches separately, and finally adapting the VAE decoder to the compensated latent representation.
The VAE encoder remains frozen throughout the entire training pipeline.
\Cref{tab:resarc_training_schedule} summarizes the optimization settings for all stages, which are detailed in the following subsections.

\paragraph{Training data.}
We filter COYO-700M~\citep{byeon2022coyo} to retain images with a shorter side greater than $1,024$ pixels, a LAION aesthetic score of at least $5.0$, and a watermark score no greater than $0.1$.
We further remove samples rejected by caption or domain filters, as well as images whose available NSFW score from either OpenNSFW2 or GantMan exceeds $0.1$.
Using PP-OCRv5 mobile text detection~\citep{cui2025paddleocr} with a confidence threshold of $0.5$, we discard images if the total detected text area exceeds $5\%$ of the image, any single text region exceeds $3.5\%$, or at least $20$ text regions are detected.

\paragraph{Selecting prefix depth.}
For residual-branch training and residual-aware VAE decoder adaptation, we use prefix depths $k\in\{5,6,7,8,9,10\}$.
We follow a balanced deterministic schedule over prefix depths across training steps.

\subsection{Autoregressive Prior Fine-Tuning}
\label{app:training_ar}
The Infinity-2B transformer~\citep{han2025infinity} is fine-tuned for 2,000 iterations on $1024\times1024$ images using AdamW~\citep{loshchilov2019decoupled}, with a global batch size of $64$ and a learning rate of $6\times10^{-5}$.
We optimize the autoregressive model with the cross-entropy objective
\begin{equation}
\mathcal L_{\mathrm{AR}}
=
-\frac{1}{|\Omega|}
\sum_{(i,d)\in\Omega}
\left[
b_{i,d}\log p_{i,d}
+
(1-b_{i,d})\log(1-p_{i,d})
\right],
\label{eq:appendix_ar_loss}
\end{equation}
where $\Omega$ indexes valid token dimensions and $p_{i,d}$ denotes the predicted probability associated with $b_{i,d}$, the $d$-th dimension of token $i$, conditioned on the preceding scales and text.
The fine-tuned autoregressive model is then frozen for all subsequent training stages.

\subsection{Mixed-Latent VAE Decoder Fine-Tuning}
\label{app:training_preparation}
We further fine-tune the VAE decoder on our compression training data.
Recovering the quantization residual shifts the decoder input from the quantized latent $h_q$ toward the continuous latent $h$.
To adapt the decoder to this shift, we expose it to both latent distributions during training.
Specifically, for each image, we sample $z=h$ with probability $0.5$ and $z=h_q$ otherwise, and reconstruct
$x_0=\mathcal D_0(z)$.
This balanced sampling improves the decoder's compatibility with the compensated latents used in ResARC without over-specializing to either latent distribution.
We then optimize the VAE decoder under this mixed-latent setting for $35,000$ steps at a resolution of $512\times512$ using the following objective:
\begin{align}
\mathcal L_{\mathrm{rec}}&=\mathcal{L}_\text{MSE}(x_0,x)+2\mathcal{L}_\text{LPIPS}(x_0,x),\label{eq:appendix_preparation_rec}\\
\mathcal L_{\mathrm{ft-dec}}&=\mathcal L_{\mathrm{rec}}+\lambda_{\mathrm{adv}}\mathcal L_{\mathrm{adv}}^G, \qquad
\mathcal L_{\mathrm{adv}}^G=-\mathbb E[\mathcal F(x_0)],
\label{eq:appendix_preparation_loss}
\end{align}
where $\mathcal{L}_\text{LPIPS}$ denotes the LPIPS-VGG loss~\citep{zhang2018lpips}, and $\mathcal L_{\mathrm{adv}}^G$ is the adversarial loss enabled after $10,000$ steps. 
The discriminator $\mathcal F$ consists of a frozen DINO feature extractor~\citep{caron2021dino} and trainable discriminator heads.
Following the adaptive weighting strategy used in perceptual autoencoder training~\citep{rombach2022high}, we set the adversarial weight $\lambda_{\mathrm{adv}}$ as
\begin{equation}
\lambda_{\mathrm{adv}}=0.5\,\operatorname{sg}\!\left[
\operatorname{clip}_{[0,10^4]}
\left(
\frac{\|\nabla_W\mathcal L_{\mathrm{rec}}\|_2}
{\|\nabla_W\mathcal L_{\mathrm{adv}}^G\|_2+10^{-4}}
\right)
\right],
\label{eq:appendix_adversarial_weight}
\end{equation}
where $W$ denotes the weights of the final convolutional layer of the decoder and $\operatorname{sg}(\cdot)$ denotes the stop-gradient operator.
Starting from step $10,000$, the discriminator heads are jointly optimized using the hinge loss
\begin{equation}
\mathcal L_{\mathcal F}=\tfrac12\mathbb E\left[\max(0,1-\mathcal F(x))+\max(0,1+\mathcal F(\operatorname{sg}(x_0)))
\right].
\label{eq:appendix_discriminator_loss}
\end{equation}
The discriminator heads are optimized with AdamW using a learning rate of $5\times10^{-5}$, $\beta=(0.5,0.9)$, and zero weight decay.

\subsection{Quantization Residual Generator Training}
\label{app:training_quant}
We train the Quantization Residual Generator in two stages: flow-matching pretraining and image-space perceptual refinement.
The conditional DiT consists of $8$ blocks with width $640$, $8$ attention heads, and a latent patch size of $2\times2$. 
It predicts 32 residual channels conditioned on a 2048-channel autoregressive decoding feature $c_{\mathrm{quant}}$ from the final-scale hidden features of the penultimate layer of the autoregressive model, without condition dropout or additional text conditioning.
A fixed normalization is applied to the quantization residual before training and inverted after sampling.

\paragraph{Flow-matching pretraining.}
We first train the conditional DiT for $3,000$ steps using the
flow-matching objective defined in~\Cref{sec:quant_residual_generation}.
The timestep is clipped to \([10^{-5},1-10^{-5}]\) for numerical stability.

\paragraph{Image-space perceptual refinement.}
Flow-matching pretraining directly supervises the latent residual but does not optimize for the perceptual quality of the final reconstruction.
We therefore further train the generator for $2,000$ steps with image-space perceptual supervision.
Specifically, with $\hat{x}_q^{(k)}=\mathcal D_0(\hat h_q^{(k)}+\hat r_q^{(k)})$, we optimize the generator using the training objective 
\begin{equation}
\mathcal L_q^{\mathrm{image}}
=
0.25\mathcal L_{\mathrm{FM}}
+
\lambda_{\mathrm{LPIPS}}(s)\,
\mathcal L_{\mathrm{LPIPS}}(\hat{x}_q^{(k)},x).
\label{eq:appendix_quant_image_loss}
\end{equation}
The perceptual weight $\lambda_{\mathrm{LPIPS}}(s)$ increases linearly from $0$ to $1$ during the first $300$ optimization steps of this stage and is maintained at $1$ thereafter.
This gradual ramp-up introduces perceptual supervision smoothly while retaining the flow-matching objective.

\begin{table}[t]
\centering
\small
\setlength{\abovecaptionskip}{2pt}
\caption{\protect\raggedright\textbf{Training hyperparameters across stages.}}
\label{tab:resarc_training_schedule}
\setlength{\tabcolsep}{4pt}
\renewcommand{\arraystretch}{1.12}

\begin{tabularx}{0.85\linewidth}{>{\raggedright\arraybackslash}Xccc}
\toprule
\rowcolor{black!4}
\textbf{Stage} & \textbf{Steps} & \textbf{Global batch} & \textbf{LR} \\
\midrule

\rowcolor{rowblue!40}
\textbf{1. Autoregressive prior fine-tuning}
& $2,000$ & $64$ & $6\times10^{-5}$ \\

\addlinespace[2pt]
\rowcolor{rowblue!40}
\textbf{2. Mixed-latent VAE decoder fine-tuning}
& $35,000$ & $16$ & $2\times10^{-5}$ \\

\addlinespace[2pt]
\rowcolor{rowblue!40}
\multicolumn{4}{l}{\textbf{3. Quantization residual generator training}} \\
\hspace*{1em}Flow-matching pretraining
& $3,000$ & $64$ & $1\times10^{-4}$ \\
\hspace*{1em}Image-space perceptual refinement
& $2,000$ & $32$ & $1\times10^{-5}$ \\

\addlinespace[2pt]
\rowcolor{rowblue!40}
\multicolumn{4}{l}{\textbf{4. Generation residual codec training}} \\
\hspace*{1em}Latent-space rate-distortion training
& $5,000$ & $8$ & $1\times10^{-4}$ \\
\hspace*{1em}Image-space perceptual refinement
& $300$ & $8$ & $1\times10^{-5}$ \\
\hspace*{1em}Joint fine-tuning
& $2,000$ & $8$ & $1\times10^{-5}$ \\

\addlinespace[2pt]
\rowcolor{rowblue!40}
\textbf{5. Residual-aware VAE decoder adaptation}
& $1,000$ & $16$ & $1\times10^{-6}$ \\

\bottomrule
\end{tabularx}
\end{table}

\subsection{Generation Residual Codec Training}
\label{app:training_generation}

\paragraph{Latent-space rate-distortion training.}
We first train the complete codec for $5,000$ steps using a latent-space rate-distortion objective.
Specifically, the generation residual reconstruction is supervised with Smooth L1 loss,
\begin{equation}
\mathcal L_{\mathrm{res}}=\mathcal{L}_\text{SmoothL1}\left(\hat r_g^{(k)},r_g^{(k)}\right),
\end{equation}
where the transition parameter of the Smooth L1 loss is set to $\beta_{\mathrm{SL1}}=0.02$. 
The entropy model provides the differentiable rate estimate $R_g^{(k)}$ defined in~\Cref{eq:appendix_generation_residual_rate}.
Combining these, the overall training objective is
\begin{equation}
\mathcal L_g^{\mathrm{latent}}=R_g^{(k)}+\mathcal L_{\mathrm{res}}.
\end{equation}

\paragraph{Image-space perceptual refinement.}
Latent-space reconstruction supervision alone does not directly optimize the perceptual quality of the reconstructed image.
We therefore freeze the analysis transform and entropy model and refine only the synthesis transform for $300$ steps using
\begin{equation}
\mathcal L_g^{\mathrm{synthesis}}
=
0.05\mathcal L_{\mathrm{res}}
+
0.8\mathcal L_{\mathrm{DISTS}}\left(\hat{x}_g^{(k)},x\right).
\end{equation}
This stage encourages the decoded generation residual to better preserve structural and textural information that is important for perceptual reconstruction.

\paragraph{Joint fine-tuning.}
Finally, we jointly fine-tune the complete Generation Residual Codec for $2,000$ steps with rate, latent reconstruction, and perceptual supervision:
\begin{equation}
\mathcal L_g^{\mathrm{full}}=R_g^{(k)}+0.05\mathcal L_{\mathrm{res}}+0.8\mathcal L_{\mathrm{DISTS}}\left(\hat{x}_g^{(k)},x\right).
\end{equation}

\subsection{Residual-Aware VAE Decoder Adaptation}
\label{app:training_adaptation}

After training the two residual branches, we freeze them and adapt
only the VAE decoder for $1,000$ iterations using the compensated
latent representation
\begin{equation}
\hat h_c^{(k)}=\hat h_q^{(k)}+\hat r_q^{(k)}+\hat r_g^{(k)}.
\end{equation}
Given the reconstruction $\hat x^{(k)}=\tilde{\mathcal D}(\hat h_c^{(k)})$, we optimize
\begin{equation}
\mathcal L_{\mathrm{adapt}}=0.05\mathcal L_{\mathrm{Charb}}+
\mathcal L_{\mathrm{LPIPS}}(\hat x^{(k)},x)+0.8\mathcal L_{\mathrm{DISTS}}(\hat x^{(k)},x),
\end{equation}
where $\mathcal L_{\mathrm{Charb}}$ denotes the Charbonnier reconstruction loss~\citep{charbonnier1994two} with $\epsilon_{\mathrm{Charb}}=10^{-3}$.
This final stage adapts the decoder to the latent distribution produced after the quantization and generation residual compensation.

\subsection{Optimization Settings}
\label{app:training_optimization}
All training stages use AdamW~\citep{loshchilov2019decoupled} with linear warm-up followed by cosine learning-rate decay and gradient-norm clipping at $1$.
Training the quantization residual generator uses $\beta=(0.9,0.999)$, zero weight decay, and a final learning-rate ratio of $0.2$, while training the generation residual codec uses the same $\beta$, a weight decay of $10^{-4}$, and a final learning-rate ratio of $0.1$.
Both VAE decoder fine-tuning and residual-aware decoder adaptation use $\beta=(0.9,0.95)$ with zero weight decay.
Stage-specific training iterations, global batch sizes, and learning rates are summarized in~\Cref{tab:resarc_training_schedule}.

\section{Bitstream and Rate Accounting}
\label{app:rate_decomposition}
The two residual branches incur different rate costs: quantization residual generation requires no additional payload, whereas generation residual coding introduces an additional transmitted bitstream.
We report the measured bitrate contributions of the text condition, token prefix, and generation-residual bitstream.
Specifically, when compressing an $H\times W$ image, the total bitrate is
\begin{equation}
R_{\mathrm{total}}^{(k)}=\frac{B_{\mathrm{text}}+B_{\mathrm{prefix}}^{(k)}+B_g^{(k)}}{HW},
\label{eq:appendix_rate_accounting}
\end{equation}
where $B_{\mathrm{text}}$, $B_{\mathrm{prefix}}^{(k)}$, and $B_g^{(k)}$ denote the measured bit lengths of the text condition, transmitted token prefix, and generation-residual bitstream, respectively.
The generation residual latent is entropy-coded using byte-rANS~\citep{duda2013ans}, and $B_g^{(k)}$ is measured from the complete encoded residual stream.
The text rate is computed from its UTF-8 byte length.

As reported in~\Cref{tab:resarc_rate_breakdown}, the generation-residual bitstream occupies only $1.98\times10^{-4}$ - $3.14\times10^{-4}$ bpp across the evaluated prefix depths.
Its fraction of the total bitrate decreases from $6.64\%$ at $k=5$ to $0.36\%$ at $k=10$, as the transmitted token prefix increasingly dominates the overall rate.
\Cref{fig:resarc_generation_residual_share} further visualizes the average bitrate composition across the prefix depths on DIV2K.
These results show that generation residual coding incurs only a small fraction of the overall transmitted rate.

\begin{figure}[t]
\centering

\includegraphics[width=1.0\linewidth]{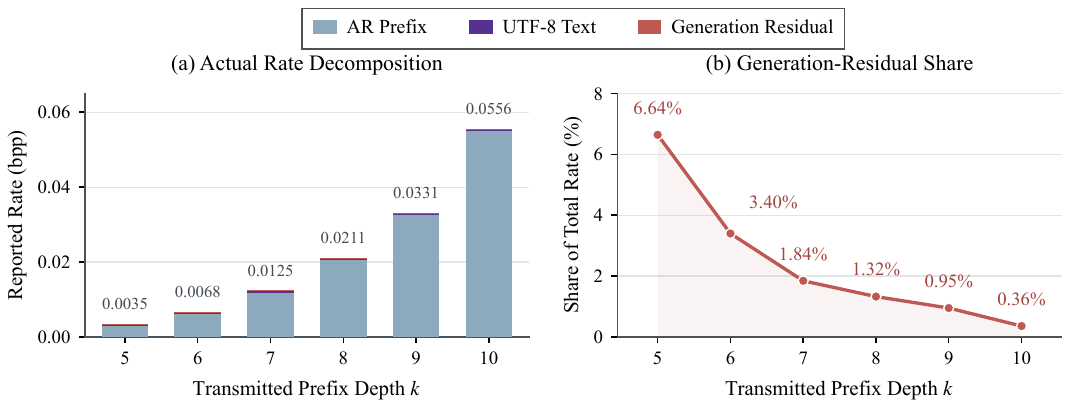}

\caption{\protect\raggedright \textbf{Rate allocation on DIV2K.}
The left panel shows the bitrate of each transmitted component across prefix depths, while the right panel shows the fraction of the total bitrate contributed by the generation-residual stream.}

\label{fig:resarc_generation_residual_share}
\end{figure}

\begin{table}[t]
\centering
\small
\caption{\protect\raggedright \textbf{Rate decomposition on DIV2K.} 
Rates are reported in bpp and averaged over the DIV2K validation set.}
\label{tab:resarc_rate_breakdown}
\setlength{\tabcolsep}{15pt}
\renewcommand{\arraystretch}{1.08}
\begin{tabular}{@{}ccccc@{}}
\toprule
$k$ & Text & AR Prefix & Gen. Residual & Total \\
\midrule
5  & $3.75\times10^{-4}$ & $2.85\times10^{-3}$ & $2.30\times10^{-4}$ & $3.46\times10^{-3}$ \\
6  & $3.75\times10^{-4}$ & $6.16\times10^{-3}$ & $2.30\times10^{-4}$ & $6.76\times10^{-3}$ \\
7  & $3.75\times10^{-4}$ & $1.18\times10^{-2}$ & $2.29\times10^{-4}$ & $1.25\times10^{-2}$ \\
8  & $3.75\times10^{-4}$ & $2.05\times10^{-2}$ & $2.79\times10^{-4}$ & $2.11\times10^{-2}$ \\
9  & $3.75\times10^{-4}$ & $3.24\times10^{-2}$ & $3.14\times10^{-4}$ & $3.31\times10^{-2}$ \\
10 & $3.75\times10^{-4}$ & $5.51\times10^{-2}$ & $1.98\times10^{-4}$ & $5.56\times10^{-2}$ \\
\bottomrule
\end{tabular}
\end{table}

\section{Evaluation Protocol}
\label{app:evaluation_protocol}
We evaluate saved 8-bit RGB reconstructions at a fixed resolution of $1024\times1024$ on the DIV2K validation set and CLIC2020 test set, containing 100 and 428 images, respectively.
Reconstruction quality is evaluated from three complementary perspectives: distortion-oriented fidelity, perceptual similarity, and distributional fidelity.

\paragraph{Evaluation settings.}
We evaluate transmitted prefix depths $k\in\{5,6,7,8,9,10\}$ within the 13-scale token hierarchy.
Autoregressive suffix generation uses multinomial sampling with temperature $0.5$, top-2, top-$p=0.97$, and classifier-free guidance scale $3.0$ applied to the logits.
The Quantization Residual Generator uses $N=4$ sampling steps in~\Cref{alg:resarc_quant_sampler}, with unit noise temperature and no classifier-free guidance.

\paragraph{Distortion-oriented fidelity metrics.}
We report PSNR and MS-SSIM~\citep{wang2003multiscale} to evaluate conventional reconstruction fidelity.
Both metrics are computed on full-resolution RGB images in $[0,1]$.
PSNR is averaged over images in dB, while MS-SSIM uses the default five-scale setting of \texttt{pytorch-msssim}.

\paragraph{Perceptual similarity.}
We report LPIPS~\citep{zhang2018lpips}, DISTS~\citep{ding2020dists}, and CLIP image-to-image similarity~\citep{radford2021clip} to measure perceptual similarity between each reconstruction and its corresponding original image.
LPIPS uses AlexNet v0.1 with inputs scaled to $[-1,1]$, while DISTS is evaluated on images scaled to $[0,1]$; both are computed at the original resolution without resizing.
CLIP image-to-image similarity is computed as $100$ times the cosine similarity between $\ell_2$-normalized ViT-B/32 image embeddings using standard CLIP preprocessing.

\paragraph{Distributional fidelity.}
We evaluate the fidelity between the distributions of reconstructed and source images using FID~\citep{heusel2017fid}, KID~\citep{binkowski2018kid}, CMMD~\citep{jayasumana2024cmmd}, and FD-DINOv2~\citep{oquab2023dinov2}.
FID and KID are computed with \texttt{torchmetrics} using 2048-dimensional Inception features.
To obtain sufficient samples for distribution estimation, each $1024\times1024$ image is partitioned into two $256\times256$ patch grids with offsets $(0,0)$ and $(128,128)$ and stride $256$, yielding 25 valid patches per image.
This results in 2,500 patches for DIV2K and 10,700 patches for CLIC2020.
FID is computed from feature means and covariances, while KID uses the polynomial kernel $\kappa(u,v)=(u^\top v/2048+1)^3$ and averages 100 estimates, each computed from 1,000 randomly sampled feature vectors with a fixed random seed.
CMMD is computed from one $\ell_2$-normalized CLIP ViT-L/14@336 embedding per image using an RBF kernel with $\sigma=10$ and the diagonal-inclusive MMD estimator scaled by $1000$.
FD-DINOv2 uses one 1024-dimensional DINOv2 ViT-L/14 CLS embedding~\citep{oquab2023dinov2} per image without $\ell_2$ normalization, with inputs resized to $224\times224$ and normalized using ImageNet statistics.

\par
\Needspace{18\baselineskip}
\begin{wraptable}{r}{0.58\textwidth}
\centering
\small
\caption{
\textbf{ResARC KID on DIV2K.}
All KID entries are in $10^{-4}$ units.}
\label{tab:div2k_raw_kid}

\setlength{\tabcolsep}{5pt}
\renewcommand{\arraystretch}{1.08}

\begin{tabular}{@{}cccc@{}}
\toprule
$k$ & bpp & KID mean $\pm$ std & 10-seed range \\
\midrule
5  & 0.00346 & $5.477 \pm 1.056$  & - \\
6  & 0.00676 & $3.177 \pm 0.946$  & - \\
7  & 0.01245 & $1.380 \pm 0.830$  & - \\
8  & 0.02110 & $0.675 \pm 0.854$  & $[0.487,\,0.835]$ \\
9  & 0.03312 & $-0.397 \pm 0.846$ & $[-0.501,\,-0.243]$ \\
10 & 0.05562 & $-1.140 \pm 0.782$ & $[-1.191,\,-0.903]$ \\
\bottomrule
\end{tabular}
\end{wraptable}

\paragraph{Signed KID estimates.}
\label{app:negative_kid}
\Cref{tab:div2k_raw_kid} reports the raw KID values of ResARC on DIV2K across different prefix depths.
The reported $\mathtt{std}$ is computed across the 100 random subsets used in the main KID evaluation.
For $k=8,9,10$, we additionally repeat the evaluation with ten random seeds and report the range of the resulting mean KID values.
As $k$ increases, KID consistently decreases and becomes negative at $k=9$ and $k=10$.
Although the population KID is non-negative, its unbiased finite-sample estimator may take negative values due to sampling variance~\citep{binkowski2018kid,gretton2012kernel}.
For BD-rate computation, we retain the raw signed KID values.
For visualization only, values below $2^{-20}$ are clipped to $2^{-20}$ in the logarithmic KID plot in~\Cref{fig:resarc_all_baselines}.

\setlength{\textfloatsep}{9pt plus 2pt minus 2pt}
\setlength{\intextsep}{8pt plus 2pt minus 2pt}
\setlength{\floatsep}{8pt plus 2pt minus 2pt}
\setlength{\abovecaptionskip}{4pt}
\setlength{\belowcaptionskip}{3pt}
\setcounter{topnumber}{3}
\setcounter{bottomnumber}{2}
\setcounter{totalnumber}{5}
\setcounter{dbltopnumber}{3}
\renewcommand{\topfraction}{0.95}
\renewcommand{\bottomfraction}{0.90}
\renewcommand{\textfraction}{0.05}
\renewcommand{\floatpagefraction}{0.80}
\renewcommand{\dbltopfraction}{0.95}
\renewcommand{\dblfloatpagefraction}{0.80}
\makeatletter
\setlength{\@fptop}{0pt}
\setlength{\@fpsep}{10pt plus 2pt minus 2pt}
\setlength{\@fpbot}{0pt plus 1fil}
\makeatother

\section{Additional Rate-Quality Assessment}
\label{app:additional_rate_quality}

\subsection{Additional Rate-Quality Curves}
\label{app:additional_rate_curves}

We additionally report PSNR, MS-SSIM~\citep{wang2003multiscale}, and CLIP image-to-image similarity~\citep{radford2021clip} on the DIV2K validation and CLIC2020 test sets.
As shown in~\Cref{fig:appendix_psnr_msssim_clip}, ResARC achieves competitive CLIP similarity at low bitrates, while maintaining comparable PSNR and MS-SSIM performance.
These results complement the main perceptual evaluation in~\Cref{fig:resarc_all_baselines} by characterizing pixel-level and structural reconstruction fidelity.

\begin{figure}[t]
\centering

\includegraphics[width=0.7\linewidth]{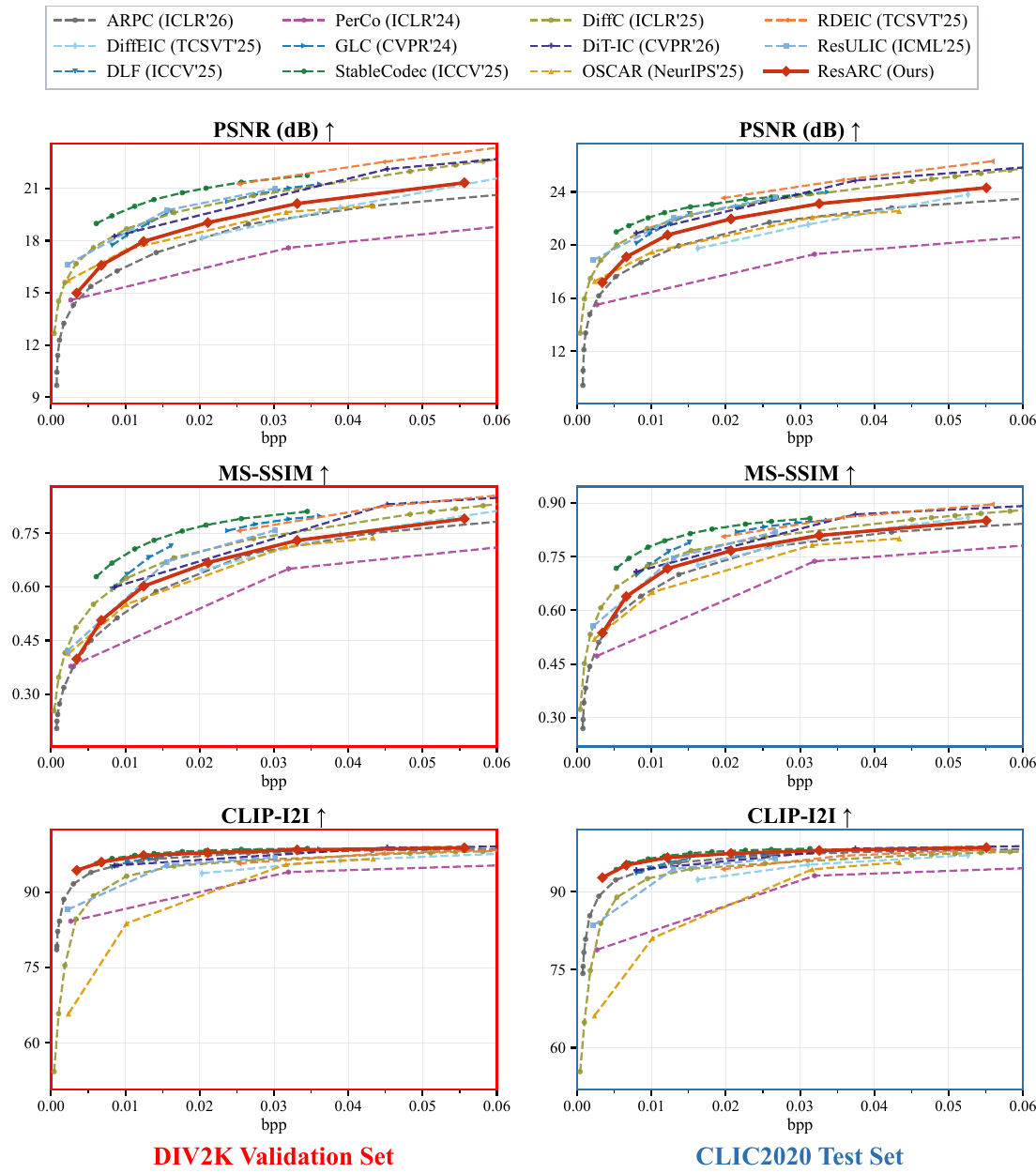}
\caption{\protect\raggedright \textbf{Additional rate-quality comparison with leading generative codecs on the DIV2K validation and CLIC2020 test sets.}}
\label{fig:appendix_psnr_msssim_clip}
\end{figure}

\subsection{BD-rate Comparison}
\label{app:bd_rate_comparison}

\Cref{tab:all_codecs_bd_arpc_extended} summarizes the signed BD-rate results across the six perceptual quality metrics.
Using ARPC~\citep{zhang2026arpc} as the reference, ResARC achieves the lowest reported BD-rates under DISTS and all four distributional fidelity metrics on both datasets, while remaining competitive under LPIPS.
Specifically, compared with the autoregressive codec ARPC, ResARC achieves BD-rate reductions of $42.94\%$ and $50.98\%$ on DIV2K under DISTS and FID, respectively, and $46.51\%$ and $51.22\%$ on CLIC2020.
ResARC also exhibits strong performance under KID, CMMD, and FD-DINOv2 while remaining competitive under LPIPS.
Compared with diffusion-based codecs such as StableCodec, ResARC achieves lower DISTS and better overall distributional fidelity at comparable bitrates.
These results further validate the effectiveness of explicitly compensating for both residuals in autoregressive generative codecs.

\begin{table}[!t]
\centering
\small

\definecolor{BDPink}{HTML}{FF1493}
\definecolor{BDBlue}{HTML}{00AEEF}
\definecolor{BDRow}{HTML}{E5F3FA}
\definecolor{BDVenue}{HTML}{555555}
\providecommand{\bdfirstfinal}[2]{}
\providecommand{\bdsecondfinal}[2]{}
\providecommand{\bddatasetfinal}[1]{}

\renewcommand{\bdfirstfinal}[2]{\textcolor{BDPink}{$\mathbf{#1}$}}
\renewcommand{\bdsecondfinal}[2]{\textcolor{BDBlue}{$\mathbf{#1}$}}
\renewcommand{\bddatasetfinal}[1]{{\fontsize{8}{9}\selectfont\bfseries #1}}

\providecommand{\bdmethodnowrap}[3]{}
\renewcommand{\bdmethodnowrap}[3]{\mbox{\fontsize{8.5}{10}\selectfont
    #1~{\color{BDVenue}(#2)}~\citep{#3}}}

\setlength{\abovecaptionskip}{2pt}
\setlength{\tabcolsep}{1.5pt}
\renewcommand{\arraystretch}{1.12}

\caption{\textbf{Signed BD-rate (\%) relative to ARPC.}
Bold \textcolor{BDPink}{pink} and \textcolor{BDBlue}{blue} denote
the lowest and second-lowest reported values for each metric and dataset.}
\label{tab:all_codecs_bd_arpc_extended}

\begin{tabularx}{\linewidth}{
    >{\raggedright\arraybackslash}X
    *{6}{>{\centering\arraybackslash}p{0.091\linewidth}}
}
\toprule
\multirow{2}{*}{\textbf{Methods}}
& \multicolumn{2}{c}{\textbf{LPIPS} $\downarrow$}
& \multicolumn{2}{c}{\textbf{DISTS} $\downarrow$}
& \multicolumn{2}{c}{\textbf{FID} $\downarrow$} \\
\cmidrule(lr){2-3}\cmidrule(lr){4-5}\cmidrule(lr){6-7}
& \bddatasetfinal{DIV2K}
& \bddatasetfinal{CLIC2020}
& \bddatasetfinal{DIV2K}
& \bddatasetfinal{CLIC2020}
& \bddatasetfinal{DIV2K}
& \bddatasetfinal{CLIC2020} \\
\midrule

\bdmethodnowrap{ARPC}{ICLR'26}{zhang2026arpc}
& 0.00 & 0.00
& 0.00 & 0.00
& 0.00 & 0.00 \\

\bdmethodnowrap{DiffEIC}{TCSVT'25}{li2025diffeic}
& +106.16 & +104.11
& +520.97 & +350.87
& +423.82 & +466.27 \\

\bdmethodnowrap{DLF}{ICCV'25}{xue2025dlf}
& -32.50 & -31.46
& +22.63 & +3.75
& +42.75 & +42.42 \\

\bdmethodnowrap{PerCo}{ICLR'24}{careil2024perco}
& +190.35 & +243.61
& +639.33 & +567.67
& +770.14 & +1207.55 \\

\bdmethodnowrap{GLC}{CVPR'24}{jia2024glc}
& -34.72 & -33.83
& +43.01 & +24.56
& +76.14 & +95.65 \\

\bdmethodnowrap{StableCodec}{ICCV'25}{zhang2025stablecodec}
& \bdsecondfinal{-39.81}{} & \bdsecondfinal{-39.78}{}
& +1.83 & -7.92
& \bdsecondfinal{-16.06}{} & \bdsecondfinal{-12.71}{} \\

\bdmethodnowrap{DiffC}{ICLR'25}{vonderfecht2025diffc}
& +31.40 & +17.23
& +359.73 & +161.03
& +417.41 & +293.57 \\

\bdmethodnowrap{DiT-IC}{CVPR'26}{shi2026ditic}
& \bdfirstfinal{-55.33}{} & \bdfirstfinal{-57.20}{}
& \bdsecondfinal{-8.88}{} & \bdsecondfinal{-19.24}{}
& +99.10 & +132.96 \\

\bdmethodnowrap{OSCAR}{NeurIPS'25}{guo2025oscar}
& +144.50 & +178.24
& +224.40 & +236.37
& +751.72 & +876.05 \\

\bdmethodnowrap{RDEIC}{TCSVT'25}{li2025rdeic}
& -7.45 & -1.43
& +387.95 & +229.43
& +272.37 & +402.97 \\

\bdmethodnowrap{ResULIC}{ICML'25}{ke2025resulic}
& +35.52 & +43.55
& +297.70 & +217.00
& +236.68 & +261.12 \\

\midrule
\rowcolor{BDRow}
\textbf{ResARC (Ours)}
& -28.88 & -31.53
& \bdfirstfinal{-42.94}{} & \bdfirstfinal{-46.51}{}
& \bdfirstfinal{-50.98}{} & \bdfirstfinal{-51.22}{} \\
\bottomrule
\end{tabularx}

\par\vspace{8pt}

\begin{tabularx}{\linewidth}{
    >{\raggedright\arraybackslash}X
    *{6}{>{\centering\arraybackslash}p{0.091\linewidth}}
}
\toprule
\multirow{2}{*}{\textbf{Methods}}
& \multicolumn{2}{c}{\textbf{KID} $\downarrow$}
& \multicolumn{2}{c}{\textbf{CMMD} $\downarrow$}
& \multicolumn{2}{c}{\textbf{FD-DINOv2} $\downarrow$} \\
\cmidrule(lr){2-3}\cmidrule(lr){4-5}\cmidrule(lr){6-7}
& \bddatasetfinal{DIV2K}
& \bddatasetfinal{CLIC2020}
& \bddatasetfinal{DIV2K}
& \bddatasetfinal{CLIC2020}
& \bddatasetfinal{DIV2K}
& \bddatasetfinal{CLIC2020} \\
\midrule

\bdmethodnowrap{ARPC}{ICLR'26}{zhang2026arpc}
& 0.00 & \bdsecondfinal{0.00}{}
& 0.00 & 0.00
& 0.00 & 0.00 \\

\bdmethodnowrap{DiffEIC}{TCSVT'25}{li2025diffeic}
& +639.93 & +736.06
& +634.54 & +39.11
& +204.14 & +136.24 \\

\bdmethodnowrap{DLF}{ICCV'25}{xue2025dlf}
& +28.85 & +55.91
& +204.20 & +115.17
& +5.94 & -1.91 \\

\bdmethodnowrap{PerCo}{ICLR'24}{careil2024perco}
& +2936.40 & +1902.80
& +1519.05 & +556.87
& +194.30 & +268.23 \\

\bdmethodnowrap{GLC}{CVPR'24}{jia2024glc}
& +164.57 & +191.31
& +91.73 & +166.07
& -21.18 & -16.97 \\

\bdmethodnowrap{StableCodec}{ICCV'25}{zhang2025stablecodec}
& \bdsecondfinal{-8.53}{} & +21.16
& \bdsecondfinal{-32.59}{} & \bdsecondfinal{-61.17}{}
& \bdsecondfinal{-31.57}{} & \bdsecondfinal{-34.50}{} \\

\bdmethodnowrap{DiffC}{ICLR'25}{vonderfecht2025diffc}
& +792.74 & +430.30
& +479.16 & +78.25
& +128.05 & +97.77 \\

\bdmethodnowrap{DiT-IC}{CVPR'26}{shi2026ditic}
& +464.09 & +353.40
& +583.35 & +153.67
& +52.70 & +26.33 \\

\bdmethodnowrap{OSCAR}{NeurIPS'25}{guo2025oscar}
& +894.84 & +1012.69
& +1857.37 & +839.67
& +460.83 & +573.09 \\

\bdmethodnowrap{RDEIC}{TCSVT'25}{li2025rdeic}
& +538.40 & +700.73
& +380.34 & +114.42
& +97.88 & +64.99 \\

\bdmethodnowrap{ResULIC}{ICML'25}{ke2025resulic}
& +801.30 & +407.62
& +312.75 & +45.41
& +47.39 & +68.55 \\

\midrule
\rowcolor{BDRow}
\textbf{ResARC (Ours)}
& \bdfirstfinal{-73.11}{} & \bdfirstfinal{-42.84}{}
& \bdfirstfinal{-57.75}{} & \bdfirstfinal{-82.74}{}
& \bdfirstfinal{-40.15}{} & \bdfirstfinal{-38.63}{} \\
\bottomrule
\end{tabularx}

\end{table}

\section{More Visual Results}
\label{app:additional_rate_results}

\subsection{Visual Effect of Prefix Depth}
\label{app:different_k}
To inspect progressive rate control, \Cref{fig:different_k_appendix} compares the same five images as the transmitted prefix depth increases from $k=5$ to $k=10$.
As more ground-truth prefix tokens are transmitted, the bitrate increases and fewer suffix scales need to be autoregressively generated, leading to progressively improved recovery of fine-grained details and image structures, including butterfly markings, mushroom textures, railing patterns, and facial details.

\begin{figure}[!htbp]
\centering

\includegraphics[pagebox=cropbox,clip,width=1.0\linewidth]{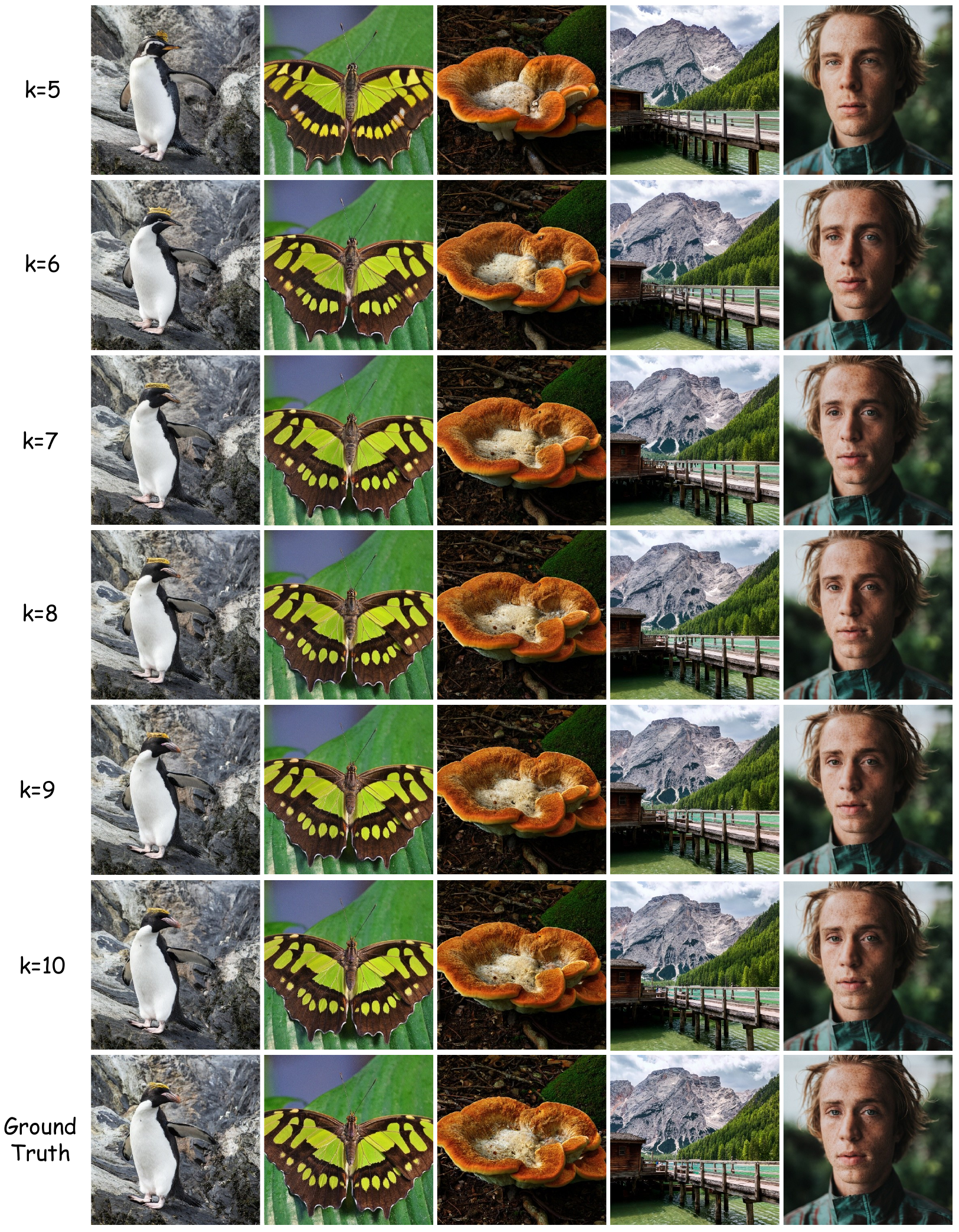}
\caption{\textbf{Progressive reconstruction across different prefix depths.}
Increasing $k$ transmits more ground-truth prefix scales and progressively improves reconstruction quality.}
\label{fig:different_k_appendix}
\end{figure}
\subsection{Additional Visual Comparisons}
\label{app:additional_visuals}

We provide additional qualitative comparisons with diffusion-based and autoregressive generative codecs in~\Cref{fig:appendix_visual1,fig:appendix_visual2}.
ResARC consistently preserves finer details and image structures, including facial details, headlight contours, glass boundaries, plate rims, and petal textures.
It also achieves the lowest full-image DISTS in every displayed example.
Together with~\Cref{fig:resarc_visual_comparison}, these results further demonstrate the effectiveness of ResARC at ultra-low bitrates.
The improvements are consistent with our residual-aware design: quantization residual generation recovers fine-grained information lost during tokenization, while generation residual compensation reduces inaccuracies introduced by autoregressive suffix generation.

\clearpage
\noindent
\begin{minipage}{\linewidth}
\centering

\includegraphics[pagebox=cropbox,width=\linewidth]{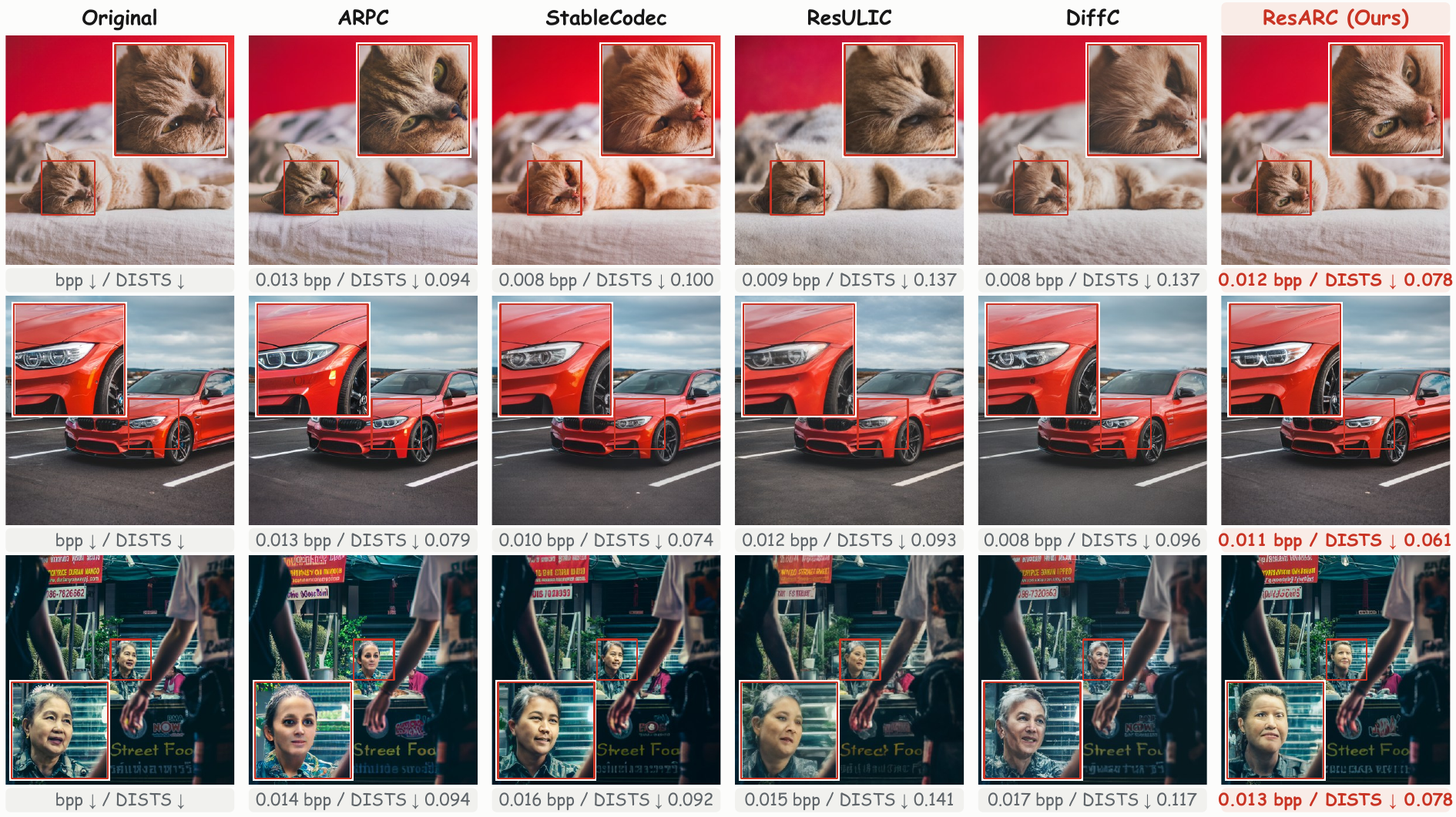}

\captionof{figure}{\protect\raggedright
\textbf{More visual comparisons on CLIC2020.} Labels report per-image bitrate (bpp) and DISTS.}

\label{fig:appendix_visual1}
\end{minipage}
\par

\par\medskip
\noindent
\begin{minipage}{\linewidth}
\centering
\includegraphics[pagebox=cropbox,width=1.0\linewidth]{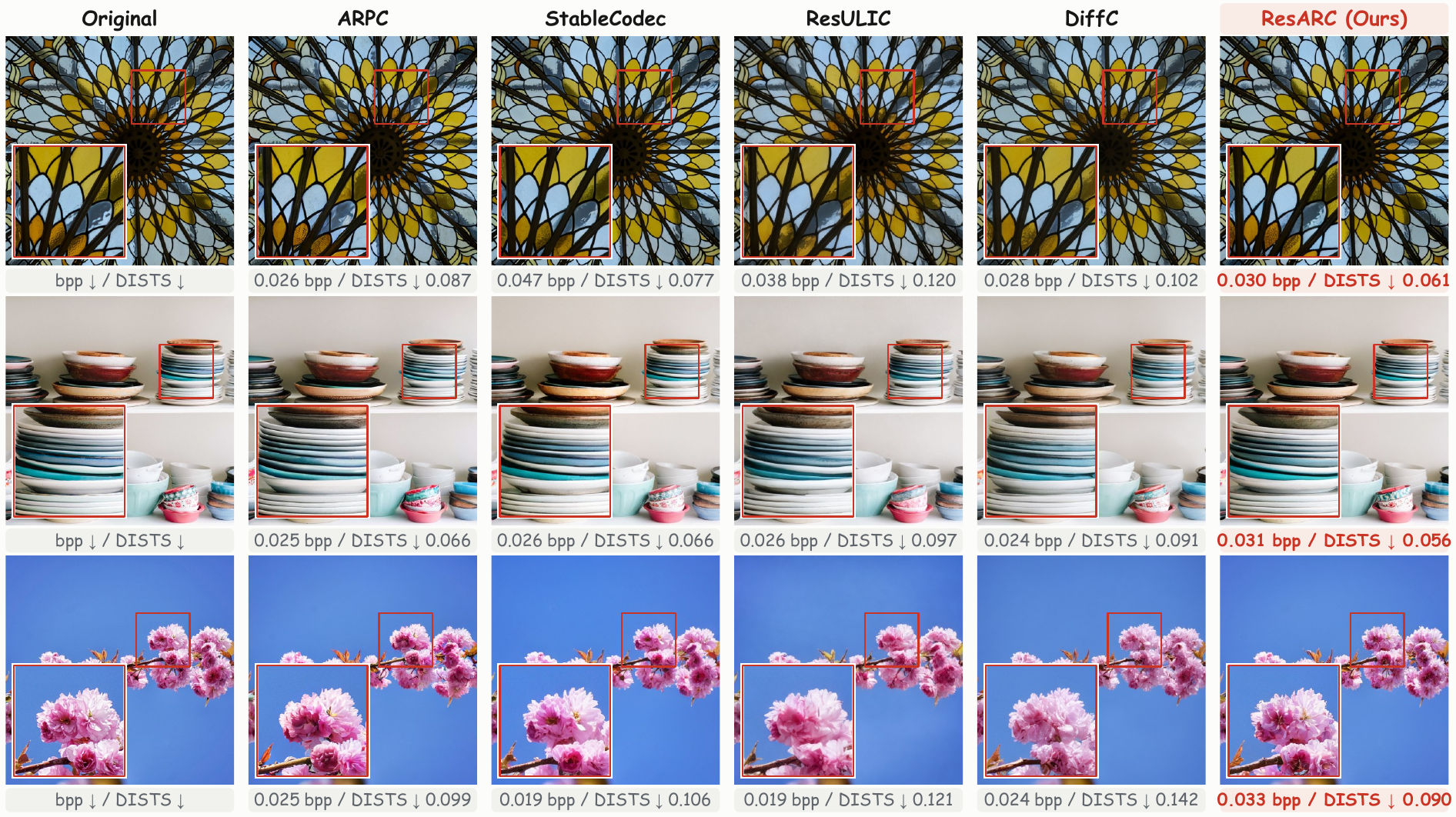}

\captionof{figure}{\protect\raggedright
\textbf{More visual comparisons on DIV2K.} Labels report per-image bitrate (bpp) and DISTS.}

\label{fig:appendix_visual2}
\end{minipage}
\par

\clearpage
\endgroup

\end{document}